\AddToHook{begindocument/end}{%
  \iclrfinalcopy
  \let\preprintmaketitle\maketitle
  \renewcommand{\maketitle}{\preprintmaketitle\lhead{Preprint}}%
}

\RequirePackage{fix-cm}
\documentclass{article} %
\usepackage{iclr2027_conference,times}

\usepackage{amsmath,amsfonts,bm}

\def\eqref#1{equation~\ref{#1}}

\def\1{\bm{1}}

\def\vone{{\bm{1}}}
\def\vmu{{\bm{\mu}}}

\def\vg{{\bm{g}}}

\def\vm{{\bm{m}}}

\def\vq{{\bm{q}}}

\def\vt{{\bm{t}}}
\def\vu{{\bm{u}}}

\def\vw{{\bm{w}}}
\def\vx{{\bm{x}}}

\def\vz{{\bm{z}}}

\def\mG{{\bm{G}}}

\def\mI{{\bm{I}}}

\def\mT{{\bm{T}}}
\def\mU{{\bm{U}}}

\def\mW{{\bm{W}}}

\def\mZ{{\bm{Z}}}

\def\mSigma{{\bm{\Sigma}}}

\DeclareMathAlphabet{\mathsfit}{\encodingdefault}{\sfdefault}{m}{sl}
\SetMathAlphabet{\mathsfit}{bold}{\encodingdefault}{\sfdefault}{bx}{n}

\DeclareMathOperator*{\argmax}{arg\,max}
\DeclareMathOperator*{\argmin}{arg\,min}

\DeclareMathOperator{\Tr}{Tr}

\usepackage{hyperref}
\usepackage{url}
\usepackage{booktabs}
\usepackage{amssymb}
\usepackage{xcolor}
\definecolor{purple}{RGB}{128,0,128}
\usepackage{colortbl}
\usepackage{graphicx}
\usepackage{tikz}
\usepackage{subcaption}
\usepackage{wrapfig}
\usetikzlibrary{arrows.meta,calc,positioning,fit,backgrounds,decorations.pathreplacing}
\definecolor{figBlue}{RGB}{39,105,166}
\definecolor{figTeal}{RGB}{26,143,126}
\definecolor{figOrange}{RGB}{218,133,47}
\definecolor{figInk}{RGB}{39,51,65}
\definecolor{figMuted}{RGB}{110,124,137}
\definecolor{figLight}{RGB}{243,246,249}

\title{Query-Time Calibration of CLIP Features for Open-Vocabulary 3D Segmentation}

\author{Dominik Engel\textsuperscript{1}\thanks{shared first-authorship},
\quad Yousef Rajeh\textsuperscript{1}\footnotemark[1],
\quad Leon Sick\textsuperscript{2}
\quad Ivan Viola\textsuperscript{1} 
\\
\textsuperscript{1} King Abdullah University of Science and Technology (KAUST)
\quad \textsuperscript{2} Ulm University
\\
}

\iclrfinalcopy %

\begin{document}

\maketitle

\begin{abstract}
Lifting vision-language features such as CLIP into 3D Gaussian reconstructions enables open-vocabulary segmentation through natural-language queries. Although CLIP represents a broad range of semantic concepts, the lifted features of individual scenes can exhibit strong directional concentration, while raw cosine similarity does not account for the scene-specific feature distribution. We show that substantial segmentation gains can be obtained by changing how a fixed lifted feature field is compared with text. Our method partially subtracts a shared scene direction from visual features and text prototypes and whitens text prototypes using a fixed reference vocabulary of generic terms. It then uses soft clustering for multiclass prediction and CSLS with a relevance margin against the reference vocabulary for single-query prediction. Beyond reducing directional concentration, scene centering improves class-relative feature separation in every evaluated scene. Across seven lifting methods and four datasets, the resulting query-time calibration consistently improves mean intersection over union in both multiclass and single-query segmentation. These gains require no retraining, source-image re-encoding, or feature re-lifting, demonstrating the value of scene-conditioned comparison for making better use of existing lifted representations.
\end{abstract}

\begin{figure}[htbp]
  \centering
  \input{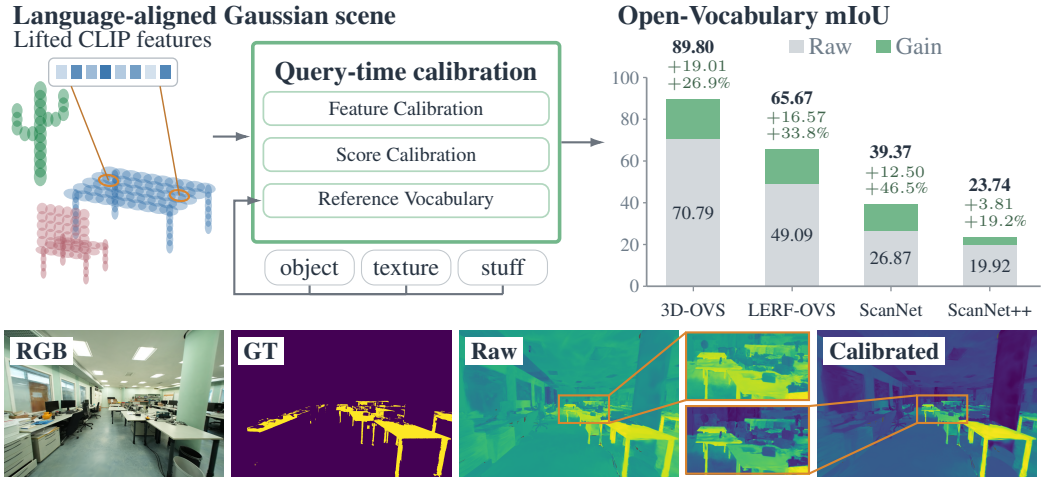}
  \caption{\textbf{Method Overview.} 
  Starting from a Gaussian scene with lifted CLIP features, we calibrate the scene and text query features using centering and text-whitening, based on a reference vocabulary. Instead of plain query-scene similarities, we further calibrate the similarity scores using soft clustering for multi-class and CSLS with a relevance margin for single-query. Our calibration results in increased class-relative separation, as visualized in the similarity maps on the bottom. Our final predictions achieve substantial improvements in segmentation mIoU for both multi-class (visualized in bar chart) and single class. 
 }
  \label{fig:overview-teaser}
\end{figure}

\section{Introduction}\label{sec:intro}
High-quality 3D reconstructions provide persistent, navigable models of captured environments. Lifting vision--language features into them lets users locate and segment objects with natural-language queries rather than a closed classifier. CLIP supplies the shared image--text space~\citep{radford2021clip}, and methods like LERF~\citep{kerr2023lerf}, LangSplat~\citep{qin2024langsplat}, and Feature 3DGS~\citep{zhou2024feature3dgs} embed or distill language-aligned features into radiance fields and Gaussian representations. As reconstruction and feature lifting improve, comparing the resulting scene with text becomes an increasingly important part of open-vocabulary segmentation.

A particular reconstructed scene typically only presents a narrower semantic distribution than the broad image--text collection used to train CLIP. We find that its lifted features can share a pronounced scene-dependent direction, while raw cosine comparisons do not explicitly account for the actual distribution observed in that scene. This presents an opportunity to improve
open-vocabulary queries solely by adapting them to
a given scene at hand, rather than assuming that global embedding geometry is equally informative in every environment. We find that a substantial part of segmentation performance can be recovered at query time, through simple calibration rather than changes to reconstruction or feature lifting.

This scene-conditioned comparison should remain useful whether a user supplies a full class vocabulary or asks for a single object. Multiclass evaluation lets candidate classes compete for each region through an argmax: a region belongs to the class that scores highest among the supplied alternatives. In practice, however, a user may simply ask for a ``fire extinguisher,'' without naming everything else in the scene. The task is then not to choose a winner among known alternatives, but to decide where the target is present and reject the remaining regions. Strong multiclass performance does not by itself establish this ability, motivating separate evaluation of single-query segmentation.

We address both settings through scene-conditioned calibration built around a fixed reference bank of generic terms. Our approach partially subtracts a shared scene direction from visual features and text prototypes and fits a regularized text-whitening transform solely to a scene-centered reference bank. For multiple requested classes, the calibrated similarities guide soft clustering of the primitive features similar to TransCLIP's assignment refinement~\citep{zanella2024transclip} without graph regularization. For a single query, we apply reduced CSLS~\citep{conneau2018csls} to the query and reference-bank scores. The reference bank then supplies negative competitors and we threshold the query's margin over their strongest reference competitor. This allows regions to be rejected without requiring the names of the scene's other classes. Our method improves comparison with already-lifted features without changing the reconstructed geometry, re-encoding source images, or re-lifting features.

Centering, whitening, and neighborhood-normalized similarity have established precedents~\citep{mu2018allbuttop,su2021whitening,conneau2018csls} in other domains, as does the use of generic negatives in language relevance scoring~\citep{kerr2023lerf}. We show that these simple operations, adapted to the scene at hand, substantially improve open-vocabulary segmentation on existing lifted feature fields. The improvements hold across different lifting methods and extend to single-query segmentation, without retraining or re-lifting features.
In summary, our contributions are:
\begin{itemize}
    \item \textbf{Query-time improvements on fixed lifted fields.}
    We show that simple scene-conditioned calibration consistently improves open-vocabulary segmentation across seven lifting methods on four datasets, in both multiclass and single-query settings, without retraining, source-image re-encoding, or feature re-lifting.
    \item \textbf{Reference-bank calibration for multiclass and single-query segmentation.}
    We combine scene centering, reference-only text whitening, and scene-relative score adjustment in a query-time method. A fixed bank of generic terms defines the text transform independently of the requested class list and supplies negative competitors for single-query segmentation.
    \item \textbf{Analysis of feature geometry and calibration components.}
    We characterize directional concentration in lifted features and show that scene centering improves class-relative feature separation in every evaluated diagnostic scene. Component ablations examine how feature calibration, score adjustment, and reference vocabulary contribute to the final segmentation gains.
\end{itemize}

\section{Related Work}\label{sec:rw}
\paragraph{Language-aware representations.}
Language-aware reconstruction methods determine which semantic information is available for subsequent queries. LERF~\citep{kerr2023lerf} embeds multiscale language features in a radiance field, while LangSplat~\citep{qin2024langsplat} and Feature 3DGS~\citep{zhou2024feature3dgs} distill features into Gaussian representations. LEGaussians~\citep{shi2023legaussians} compresses language features through quantization, FMGS~\citep{zuo2024fmgs} combines foundation-model features with a multiresolution feature field, and econSG~\citep{zhang2025econsg} addresses feature consistency across views. Gaussian Grouping~\citep{ye2024gaussiangrouping}, OpenGaussian~\citep{wu2024opengaussian}, and N2F2~\citep{bhalgat2024n2f2} additionally organize instance identities or semantic scales. These approaches improve the 3D representation constructed from images.

Efficient lifting methods reduce the cost of constructing language-aware fields. Dr.~Splat~\citep{kim2025drsplat} directly registers language features, Occam's LGS~\citep{cheng2025occam} aggregates multiview observations, and Splat Feature Solver~\citep{xiong2025splatfeaturesolver} formulates lifting as a regularized inverse problem. LightSplat~\citep{bang2026lightsplat} uses compact semantic indices and 3D clustering to associate image regions efficiently. Their feature-construction strategies are complementary to post-lifting calibration, which can operate on their outputs without repeating reconstruction or lifting.

Dense language features provide the inputs to many such pipelines. OpenSeg~\citep{ghiasi2022openseg} and LSeg~\citep{li2022lseg} learn dense visual-language alignment, while MaskCLIP~\citep{zhou2022maskclip} extracts dense predictions from CLIP~\citep{radford2021clip}. SAN~\citep{xu2023san} and OVSeg~\citep{liang2023ovseg} adapt the segmentation representation through additional networks or masked-region training. SCLIP~\citep{wang2024sclip}, ClearCLIP~\citep{lan2024clearclip}, ProxyCLIP~\citep{lan2024proxyclip}, and CLIPtrase~\citep{shao2024cliptrase} instead rectify attention or token features without training. These methods are concerned with improving text and image features directly, rather than operating at query-time on already lifted fields.

Calibration also appears in dense and 3D segmentation. SCAN~\citep{liu2024scan} learns proposal calibration and contextual adaptation, while ReCLIP~\citep{wang2024reclip} learns class and spatial bias corrections. OpenScene~\citep{peng2023openscene} fuses dense multiview features, and Open3DIS~\citep{nguyen2024open3dis} associates image-guided proposals with language evidence. PLA~\citep{ding2023pla}, RegionPLC~\citep{zhang2024regionplc}, and AIDE~\citep{wang2025aide} include learned confidence mechanisms for balancing known and novel classes. Our setting instead requires a label-free correction applicable to an already available field without learning a segmentation or confidence model.

\paragraph{Feature adaptation.}
Adapting embedding geometry provides one route to improving fixed-feature comparisons. Alignment and uniformity~\citep{wang2020alignment} characterize contrastive representations, while modality-gap analysis~\citep{liang2022gap} and Two Effects, One Trigger~\citep{schrodi2024twoeffects} identify systematic differences between visual and text embeddings. All-but-the-Top~\citep{mu2018allbuttop} and isotropy analysis~\citep{rajaee2021isotropy} motivate removing dominant directions, while sentence whitening~\citep{su2021whitening} and Whitened CLIP~\citep{betser2025whitenedclip} transform covariance structure. These works directly motivate our calibration components, with scene-local statistics and the reference vocabulary determining the calibration of a  lifted field.

Other adaptations require evidence or optimization beyond stored scene features. CLIP Surgery~\citep{li2023clipsurgery} removes image-conditioned redundant features, and CalibCLIP~\citep{kang2025calibclip} modifies dominant-token processing. CoOp~\citep{zhou2022coop} and CoCoOp~\citep{zhou2022cocoop} learn prompts, TPT~\citep{shu2022tpt} and C-TPT~\citep{yoon2024ctpt} optimize prompts at test time, and Tip-Adapter~\citep{zhang2022tipadapter} uses labeled support examples. ORION~\citep{chakraborty2026orion} fine-tunes the text encoder to separate class prototypes. Their feature or prototype corrections target related errors, but their image access, labeled support, or learned parameter updates differ from our analytic transformation of frozen embeddings.

Distribution-conditioned adaptations are more direct competitors on fixed features. UMFC~\citep{liang2024umfc} estimates domain-dependent biases from unlabeled visual collections and corrects both visual and text features. PTC~\citep{ma2026ptc} builds visual prototypes from selected predictions and uses them to adjust text embeddings. Frolic~\citep{zhu2024frolic} constructs a covariance-based classifier, combines it with CLIP predictions, and iteratively corrects class priors. Applied to lifted primitives, these mechanisms compete with our scene centering and reference-text whitening by estimating corrections from the visual distribution or predicted class structure.

\paragraph{Score calibration.}
Reference-population normalization changes how fixed embeddings compete without modifying their encoders. Inverted softmax~\citep{smith2017inverted}, CSLS~\citep{conneau2018csls}, and querybank normalization~\citep{bogolin2022qb_norm} correct retrieval scores using statistics of other samples. Hubness~\citep{radovanovic2010hubs} and prediction-concentration analysis~\citep{sato2026hubness} connect such corrections to unequal neighbor or class occurrence. Our multiclass branch uses soft clustering to refine primitive assignments, while our single-query branch applies class-side CSLS offsets before computing the query-versus-reference margin. Alternative scene-score normalizations are therefore direct comparisons for the single-query score correction.

Target-distribution methods instead adapt predictions jointly across samples or classes. BBSE~\citep{lipton2018bbse} and logit adjustment~\citep{menon2021logit} connect prediction correction to class-frequency assumptions. OTTER~\citep{shin2024otter} solves an optimal-transport assignment given an estimated target label distribution, while OnZeta~\citep{qian2024onzeta} updates label statistics and visual-space class proxies online. TransCLIP~\citep{zanella2024transclip} iteratively estimates class distributions and propagates assignments over a feature graph. We adopt its soft assignment refinement without graph regularization for multiclass prediction and combine it with scene centering and reference-bank text whitening. 

Single-query segmentation additionally requires a decision between foreground and background. LERF~\citep{kerr2023lerf} and FMGS~\citep{zuo2024fmgs} compare a query against four canonical negative phrases, N2F2~\citep{bhalgat2024n2f2} uses generic phrases in multiscale feature learning, and NegLabel~\citep{jiang2024neglabel} mines negative labels for out-of-distribution detection. GOI~\citep{qu2024goi} replaces fixed-threshold selection with a query-specific semantic hyperplane optimized using a referring-expression model, while FreeCP~\citep{chen2025freecp} removes redundant candidate classes to reduce visual-language ambiguity. These approaches address negative competition, query-specific selection, or candidate-set quality.

\section{Method}\label{sec:method}
As outlined in \autoref{fig:overview-teaser}, our approach starts with a fixed Gaussian scene with lifted CLIP features and performs two steps. First we calibrate features using scene centering (Sec.~\ref{sec:method:centering}) and text-queries are whitened (Sec.~\ref{sec:method:whitening}) using our reference vocabulary (Sec.~\ref{sec:method:vocabulary-extension}). In the second step we calibrate per-query scores for both multi-class (Sec.~\ref{sec:method:multiclass}) and single-query predictions (Sec.~\ref{sec:method:single-query}).

\subsection{Scene Centering}
\label{sec:method:centering}
\begin{wrapfigure}{r}{.48\textwidth}
\vspace{-8.7mm}
\centering
\resizebox{\linewidth}{!}{%
\begin{tikzpicture}
  \path[use as bounding box] (0,0) rectangle (7,4.75);
  \tikzset{
    panel title/.style={font=\small\bfseries, text=figInk},
    step title/.style={font=\scriptsize\bfseries, text=figInk},
    sphere/.style={draw=figInk!72, line width=.75pt, fill=none},
    raw ray/.style={-{Latex[length=1.8mm,width=1.2mm]}, draw=figBlue, line width=1pt},
    corrected ray/.style={-{Latex[length=1.8mm,width=1.2mm]}, draw=figTeal, line width=1pt},
    mean ray/.style={-{Latex[length=1.7mm,width=1.15mm]}, draw=figOrange, dashed, line width=.9pt},
    flow/.style={-{Latex[length=2mm,width=1.35mm]}, draw=figInk!78, line width=.9pt},
    inset box/.style={draw=figMuted!45, fill=white, rounded corners=2pt, line width=.55pt}
  }

  \def\lambdaval{0.3}
  \def\meanangle{35}
  \coordinate (L) at (1.45,3.38);
  \coordinate (R) at (5.55,3.38);

  \fill[figLight] (L) circle[radius=1.02];
  \path[fill=figBlue!11]
    (L) -- ([shift={(9.5:1.02)}]L)
    arc[start angle=9.5,end angle=60.5,radius=1.02] -- cycle;
  \draw[sphere] (L) circle[radius=1.02];
  \draw[draw=figMuted!65, dashed, line width=.5pt]
    ($(L)+(1.02,0)$)
    arc[start angle=0,end angle=180,x radius=1.02,y radius=.23];
  \draw[draw=figInk!58, line width=.65pt]
    ($(L)+(-1.02,0)$)
    arc[start angle=180,end angle=360,x radius=1.02,y radius=.23];
  \foreach \a in {9.5,24.8,45.2,60.5}{\draw[raw ray] (L) -- ++(\a:1.02);}
  \draw[mean ray] (L) -- ++(\meanangle:.96)
    node[pos=1, above right=2pt, font=\scriptsize, text=figOrange] {$\vmu$};
  \fill[figInk] (L) circle[radius=1.3pt];
  \node[panel title] at (1.45,4.60) {Before};
  \node[font=\scriptsize, text=figBlue, inner sep=1pt] at (1.45,2.20) {input $\vz$};

  \pgfmathsetmacro{\outlo}{35+1.10*(atan2(sin(5)-\lambdaval*sin(\meanangle),cos(5)-\lambdaval*cos(\meanangle))-35)}
  \pgfmathsetmacro{\outhi}{35+1.10*(atan2(sin(65)-\lambdaval*sin(\meanangle),cos(65)-\lambdaval*cos(\meanangle))-35)}
  \fill[figLight] (R) circle[radius=1.02];
  \path[fill=figTeal!12]
    (R) -- ([shift={(\outlo:1.02)}]R)
    arc[start angle=\outlo,end angle=\outhi,radius=1.02] -- cycle;
  \draw[sphere] (R) circle[radius=1.02];
  \draw[draw=figMuted!65, dashed, line width=.5pt]
    ($(R)+(1.02,0)$)
    arc[start angle=0,end angle=180,x radius=1.02,y radius=.23];
  \draw[draw=figInk!58, line width=.65pt]
    ($(R)+(-1.02,0)$)
    arc[start angle=180,end angle=360,x radius=1.02,y radius=.23];
  \foreach \a in {5,23,47,65}{
    \pgfmathsetmacro{\correctedangle}{35+1.10*(atan2(sin(\a)-\lambdaval*sin(\meanangle),cos(\a)-\lambdaval*cos(\meanangle))-35)}
    \draw[corrected ray] (R) -- ++(\correctedangle:1.02);
  }
  \draw[mean ray] (R) -- ++(\meanangle:.96)
    node[pos=1, above right=2pt, font=\scriptsize, text=figOrange] {$\vmu$};
  \fill[figInk] (R) circle[radius=1.3pt];
  \node[panel title] at (5.55,4.60) {After};
  \node[font=\scriptsize, text=figTeal, inner sep=1pt] at (5.55,2.20) {$C_{\lambda}(\vz)$};

  \draw[flow] (2.63,3.38) -- (4.37,3.38);
  \node[font=\scriptsize, text=figInk, fill=white, inner sep=1.2pt] at (3.50,3.72)
    {Eq.~\ref{eq:scene-centering}};

  \node[inset box, minimum width=6.70cm, minimum height=1.68cm] at (3.50,.98) {};
  \draw[draw=figMuted!30, line width=.45pt] (3.50,.18) -- (3.50,1.70);
  \node[step title] at (1.75,1.58) {Subtract};
  \node[step title] at (5.25,1.58) {Renormalize};

  \coordinate (I) at (.85,1.00);
  \coordinate (Z) at ($(I)+(-8:1.26)$);
  \coordinate (U) at ($(Z)+({\meanangle+180}:{\lambdaval*1.26})$);
  \draw[raw ray] (I) -- (Z)
    node[pos=.68, above=1pt, font=\scriptsize, text=figBlue] {$\vz$};
  \draw[-{Latex[length=1.5mm,width=1mm]}, draw=figOrange, line width=.9pt]
    (Z) -- (U)
    node[midway, right=2pt, font=\scriptsize, text=figOrange] {$-\lambda \vmu$};
  \draw[-{Latex[length=1.5mm,width=1mm]}, draw=figTeal, line width=.9pt] (I) -- (U);
  \fill[figInk] (I) circle[radius=1.15pt];
  \fill[figTeal] (U) circle[radius=1.4pt]
    node[below right=0pt, font=\scriptsize, text=figTeal] {$\vu$};

  \pgfmathsetmacro{\demoout}{atan2(sin(-8)-\lambdaval*sin(\meanangle),cos(-8)-\lambdaval*cos(\meanangle))}
  \pgfmathsetmacro{\unorm}{sqrt(1+\lambdaval*\lambdaval-2*\lambdaval*cos(-8-\meanangle))}
  \coordinate (J) at (4.72,.72);
  \coordinate (V) at ($(J)+(\demoout:{.48*\unorm})$);
  \coordinate (N) at ($(J)+(\demoout:.48)$);
  \draw[draw=figMuted!62, dashed, line width=.55pt] (J) circle[radius=.48];
  \draw[draw=figMuted!55, densely dotted, line width=.55pt] (J) -- ++(\demoout:.63);
  \draw[draw=figTeal, line width=1pt] (J) -- (V);
  \draw[-{Latex[length=1.5mm,width=1mm]}, draw=figTeal, line width=1pt] (V) -- (N);
  \fill[figInk] (J) circle[radius=1.15pt];
  \fill[figTeal] (V) circle[radius=1.4pt]
    node[below right=0pt, font=\scriptsize, text=figTeal] {$\vu$};
  \draw[draw=figTeal, fill=white, line width=.8pt] (N) circle[radius=1.7pt]
    node[above right=2pt, font=\scriptsize, text=figTeal] {$\nu(\vu)$};
  \node[font=\scriptsize, text=figMuted] at (6.18,.35) {unit sphere};
\end{tikzpicture}%
}
\vspace{-6mm}
\caption{\textbf{Scene centering} maps blue input vectors $\vz$ to teal calibrated vectors $\vu$ by subtracting the orange mean direction $\vmu$ and renormalizing.}
\label{fig:geometry-scene-centering}
\vspace{-6mm}
\end{wrapfigure}
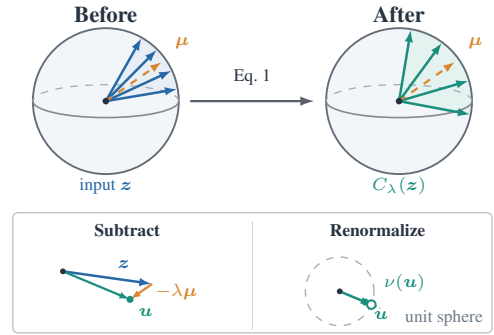

First, we calibrate scene and text features as illustrated in \autoref{fig:geometry-scene-centering}. Let $\vx_i,\vt_c\in\mathbb{R}^d$ be unit-normalized column-vector features for valid primitives $i\in\mathcal{V}$ and text prototypes $c$, respectively.
Both lie in the same CLIP space~\citep{radford2021clip} that covers much broader semantics than the scene at hand. The calibration is independent of the lifting method, provided its output features share the text encoder's CLIP space.
For a column vector $\vz\in\mathbb{R}^d$, $\nu(\vz)=\vz/\max(\lVert\vz\rVert_2,10^{-12})$ denotes normalization, for matrices $\nu$ normalizes each row.
The centering subtracts the shared scene direction, represented by the normalized mean $\vmu\in\mathbb{R}^d$:
\begin{equation}
 \vmu=\nu\!\left(\frac{1}{|\mathcal{V}|}\sum_{i\in\mathcal{V}}\vx_i\right),
 \qquad C_\lambda(\vz)=\nu(\vz-\lambda\vmu)
 \label{eq:scene-centering}
\end{equation}
with $\lambda=0.3$. For a row-stacked matrix $\mZ\in\mathbb{R}^{r\times d}$, the row-wise extension is explicitly
$C_\lambda(\mZ)=\nu(\mZ-\lambda\vone_r\vmu^\top)$.
Feature Calibration applies this centering step symmetrically, with $\widetilde{\vx}_i=C_\lambda(\vx_i)$ calibrating per-primitive features and $\widetilde{\vt}_c=C_\lambda(\vt_c)$ calibrating text query features.

\subsection{Text-query Whitening}
\label{sec:method:whitening}
Next, we whiten query features using statistics fitted only to the scene-centered reference vocabulary. For a fixed scene, this whitening transform is independent of the requested class list. We scene-center the reference features and subtract their mean.
\[
 \vg_b\in\mathbb{R}^d,\qquad
 \mG_B=
 \begin{bmatrix}
  \vg_1^\top\\
  \vdots\\
  \vg_B^\top
 \end{bmatrix}
 \in\mathbb{R}^{B\times d},\qquad \text{ with }B=45.
\]

Here $\vg_b$ is the unit text feature for entry $b$ in reference-bank $\mG_B$. Applying the centering gives
\begin{equation}
 \widetilde{\mG}_B=C_\lambda(\mG_B), \qquad
 \bar{\vg}=\frac{1}{B}\widetilde{\mG}_B^\top\vone_B,\qquad
 \mG_0=\widetilde{\mG}_B-\vone_B\bar{\vg}^{\top}.
\end{equation}

Here $\vone_B\in\mathbb{R}^B$ is the all-ones vector. With the resulting mean $\bar{\vg}\in\mathbb{R}^d$ of the centered reference bank $\widetilde{\mG}_B$ and the residual matrix $\mG_0\in\mathbb{R}^{B\times d}$, the covariance is
\begin{equation}
 \mSigma=\frac{\mG_0^\top\mG_0}{B-1}\in\mathbb{R}^{d\times d}.
 \label{eq:text-covariance}
\end{equation}

Note that only the reference bank contributes to the fitted mean and covariance. We compute the eigenvector matrix $\mU\in\mathbb{R}^{d\times d}$ to define
\[
 \begin{gathered}
 \mSigma_\rho=\mSigma+\rho\frac{\Tr(\mSigma)}{d}\mI_d,
 \qquad
 \mSigma_\rho=\mU\operatorname{diag}(\sigma_1,\ldots,\sigma_d)\mU^\top.
 \end{gathered}
\]
with $\rho=10^{-3}$ and $\widehat\sigma_j=
 \max\!\left(\sigma_j,10^{-10}\max_\ell\sigma_\ell\right) \forall j=1,\ldots,d$ being used to stabilize subsequent inverse-power scaling when computing the whitening matrix $\mW_{\alpha,B}\in\mathbb{R}^{d\times d}$:
\begin{equation}
 \mW_{\alpha,B}=
 \mU\operatorname{diag}(\widehat\sigma_1^{-\alpha},\ldots,
 \widehat\sigma_d^{-\alpha})\mU^\top,
 \qquad 
 \mathcal{W}_B(\vz)=
 \nu\!\left(\mW_{\alpha,B}(\vz-\bar{\vg})\right).
 \label{eq:text-whitening}
\end{equation}
with $\alpha=0.40$ determined experimentally.
Thus every query text feature is transformed as $\vt'_c=\mathcal{W}_B(C_\lambda(\vt_c))$. Similarly we define $\mathcal{W}_B(\mZ)=
 \nu\!\left((\mZ-\vone_r\bar{\vg}^{\top})\mW_{\alpha,B}^{\top}\right)$ for matrix $\mZ\in\mathbb{R}^{r\times d}$.
 
Only the reference bank determines $\bar{\vg}$ and $\mW_{\alpha,B}$ and requested queries do not enter these statistics. Whitening is applied only to text features, while visual features remain scene-centered as $\widetilde{\vx}_i=C_\lambda(\vx_i)$ and are not whitened. $\mathcal W_B$ defines the whitening operation we subsequently apply to queries. We use regularized inverse-power scaling slightly weaker than inverse-square-root scaling typically employed in standard whitening. While queries are not used within whitening, our query calibration remains scene-conditioned through the use of scene-centered features.

\subsection{Reference Vocabulary}
\label{sec:method:vocabulary-extension}
The fixed vocabulary represents the text-side shared concepts using generic terms. Our bank $G_B$ contains 45 reference unit text features $\vg_b$ and includes only generic concepts like ``object'', ``things'', ``texture'', and further generic scene, material, and spatial descriptors, but no class names.
Appendix~\ref{sec:auxiliary-vocabulary} lists the complete bank and its
text-encoding convention. As detailed in Sections~\ref{sec:method:multiclass} and \ref{sec:method:single-query}, we use the reference vocabulary to determine whitening statistics for text query calibration and in the single-query case as background competitors. 

To construct the vocabulary, we start from a fixed pool of 64 generic words and
phrases and successively remove words deemed unhelpful.  At each step, we test
removing each entry and remove the word that gives the largest increase in mean
mIoU on held-out calibration scenes, weighting the multi-class and single-query
tasks and the scenes equally.  We repeat this process until no removal
improves this objective, leaving 45 entries, and then fix the bank.

\begin{figure}[t]
  \centering
  \vspace{-3mm}
  \input{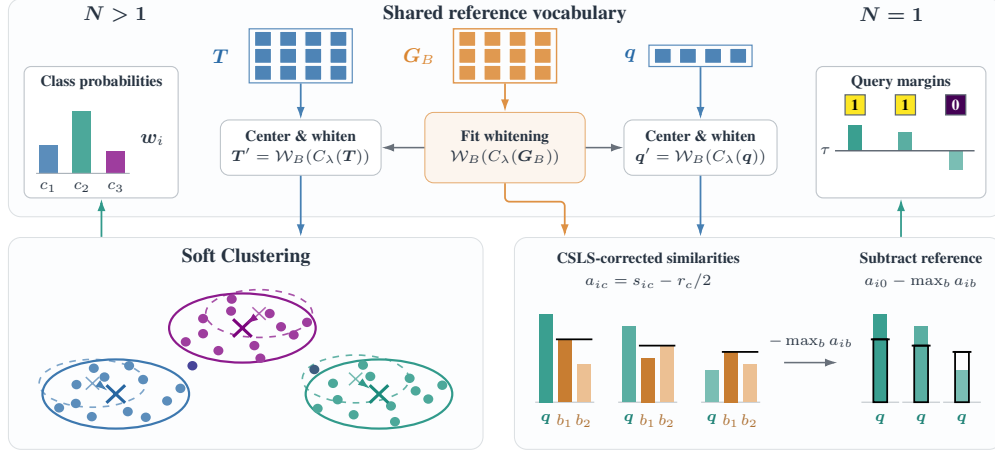}
  \vspace{-6mm}
  \caption{\textbf{Score calibration.} Starting from the \textbf{Middle:}
  First we use the reference vocabulary to determine the whitening statistics. \textbf{Left:} For multi-class queries ($N>1$) we center and whiten all text queries and perform soft clustering of the primitive features, using fixed text-derived probabilities to guide assignments and weighted top-eight visual primitives to initialize the centers. After clustering, argmax decides the class of each primitive. \textbf{Right:} For a single-class query ($N=1$) we center and whiten the query and the reference vocabulary, apply reduced CSLS and compute the margin between each primitive's corrected query score and its highest corrected reference-bank score. To make a prediction we threshold this relevance margin using $\tau$.
}
  \label{fig:score-calibration}
  \vspace{-3mm}
\end{figure}

\subsection{Multi-Class Score Calibration}
\label{sec:method:multiclass}

For $N>1$ requested classes, we adopt the soft k-means-like refinement from TransCLIP~\citep{zanella2024transclip}. Specifically, we apply the clustering to our calibrated features, but without their proposed similarity graph regularization. We alternate between soft class assignments and updates to the class centers and shared diagonal variance.
We first compare each scene-centered visual feature $\widetilde{\vx}_i$ with the calibrated text prototype $\vt'_c$ for $c=1,\ldots,N$:
\begin{equation}
 s_{ic}=\widetilde{\vx}_i^\top\vt'_c,
 \qquad
 \pi_{ic}=\frac{\exp(100s_{ic})}{\sum_{c'=1}^{N}\exp(100s_{ic'})}.
 \label{eq:multiclass-initial-probabilities}
\end{equation}
These initial class probabilities remain fixed and guide the refinement. For clustering, we initialize each class center from its eight highest-probability primitives,
weighted by their probabilities and normalized to unit length. We then
alternate between updating soft class assignments $w_{ic}$ and estimating class centers $\vm_c$
and a shared diagonal variance $v_j$ from the assigned visual features.
We update soft assignments using the initial probabilities and distances to
the current class centers, then refit the centers and shared diagonal variance:
\begin{equation}
 \begin{gathered}
 w_{ic}\propto
 \pi_{ic}\exp\!\left(
 -\frac{1}{2\beta}\sum_{j=1}^{d}
 \frac{(\widetilde{x}_{ij}-m_{cj})^2}{v_j}
 \right),
 \qquad \sum_{c=1}^{N}w_{ic}=1,
 \\
 \vm_c\leftarrow\nu\!\left(
 \frac{\sum_{i\in\mathcal V}w_{ic}\widetilde{\vx}_i}
 {\sum_{i\in\mathcal V}w_{ic}}
 \right),
 \qquad
 v_j\leftarrow\frac{1}{|\mathcal V|}
 \sum_{i\in\mathcal V}\sum_{c=1}^{N}
 w_{ic}\bigl(\widetilde{x}_{ij}-m_{cj}\bigr)^2.
 \end{gathered}
 \label{eq:multiclass-soft-assignments}
\end{equation}
For each primitive, we normalize the assignment weights by dividing the
right-hand side of the proportionality by its sum over all classes, so
$\sum_{c=1}^{N}w_{ic}=1$. We then update the class centers and variance in
that order, with $\nu$ normalizing each center to unit length.
Here $v_j$ is the shared variance in
feature dimension $j$, initialized to $1/d$. 
We use $\beta=50$ and 11 iterations, following TransCLIP's implementation~\citep{zanella2024transclip}.
Finally, We predict $\widehat y_i=\argmax_{1\leq c\leq N}w_{ic}$.

\subsection{Single-query Score Calibration}
\label{sec:method:single-query}
With one requested query $\vq$, a class argmax alone cannot reject background and we cannot perform soft clustering with a single prototype.
Instead, we calibrate the query's similarity scores using CSLS~\citep{conneau2018csls} and compare them against the most similar prototype in our reference bank $\mG_B$.

We apply the scene centering and fixed reference-bank whitening transform to both the query and the reference words. For each resulting prototype $\vt'_c$, where $c\in\{0,\ldots,B\}$ with $\vt'_0 = \vq'$, we compute its similarity to the scene features $s_{ic}=\widetilde{\vx}_i^\top\vt'_c$. Then we take the average of the top-1000 similarities per class $r_c$, which is used to correct each score: $a_{ic}=s_{ic}-\tfrac12r_c$.
This reduced CSLS correction adjusts each query's score baseline while preserving its ranking of primitives.

Let $a_{i0}$ denote the corrected query score and $a_{ib}$ the score for
reference word $b$. We subtract the strongest reference score to obtain the \emph{relevance margin}:
\begin{equation}
 R_i(\vq)=a_{i0}-\max_{1\leq b\leq B}a_{ib},
 \qquad
 \widehat M_i(\vq)=\1\!\left[R_i(\vq)\geq\tau\right].
 \label{eq:single-query-margin}
\end{equation}
The indicator $\1[\cdot]$ is one when its condition holds and zero otherwise. The margin measures whether the requested query scores above the generic reference words at that primitive. We use one threshold $\tau$ across all scenes and queries within each dataset and lifted feature set.

\section{Evaluation}\label{sec:eval}
We evaluate our calibration on its ability to increase class-relative separation, 3D open-vocabulary segmentation and present ablation studies. For the open-vocabulary segmentation we use LERF-OVS~\cite{qin2024langsplat}, 3D-OVS~\cite{liu2023weakly}, ten scenes from ScanNet~\citep{dai2017scannet} and twelve from ScanNet++~\citep{yeshwanth2023scannetpp} using mean intersection-over-union (mIoU), precision and recall. We evaluate lifted features from SFS~\citep{xiong2025splatfeaturesolver},
NormLift~\citep{zang2026normlift}, LUDVIG~\citep{marrie2026ludvig}, OpenGaussian~\citep{wu2024opengaussian}, LangSplat
~\citep{qin2024langsplat}, Occam~\citep{cheng2025occam} and VALA~\citep{wang2026vala}.
To assign ground truth points to Gaussians, we follow the respective methods' strategies for discarding low-opacity and invalidly lifted Gaussians, and assign the remaining eligible Gaussians using minimum Mahalanobis distance. Appendix~\ref{app:p3-protocol} presents more details on our evaluation protocol.

In Section~\ref{sec:eval:ov-seg} we compare against raw uncalibrated features, as well as previous work on CLIP calibration. These methods were not designed for our Gaussian-lifted 3D task, but we adopt their ideas to enable comparison. Specifically we compare to adaptations of TransCLIP~\citep{zanella2024transclip}, PTC~\citep{ma2026ptc}, UMFC~\citep{liang2024umfc} and a Frolic~\citep{zhu2024frolic} inspired baseline.

\subsection{Investigating the Effects of Feature Calibration}
We quantify the directional concentration illustrated in \autoref{fig:geometry-scene-centering}
using the von Mises--Fisher (vMF) concentration estimate $\kappa$.
Higher values indicate stronger concentration around a common direction.
As intended, subtracting the shared scene direction reduces mean $\kappa$
from 1600 to 517 across the twelve ScanNet++ SFS scenes.
The uncentered second moment averages feature outer products without subtracting
their mean, capturing both a shared direction and the spread around it.
Its leading eigenvalue's share measures the fraction of total squared feature
magnitude along the strongest axis, this falls from $76.8\%$ to $47.3\%$.
We also analyze participation rank, which summarizes how many axes effectively carry this magnitude:
$k$ equally weighted axes give a participation rank of $k$.
It rises from $1.69$ to $4.15$, indicating that the features are less dominated
by a few directions. Together, these descriptive statistics show that centering
redistributes feature magnitude across more axes, rather than merely reducing
the mean direction.

To assess semantic discrimination, we group features by their GT class and
normalize each class's average feature to obtain its mean direction.
Between-class scatter is the average squared distance between these class
mean directions: it measures how far apart the class centers are.
Within-class scatter is the average squared distance of features from their
own class's mean direction, averaged equally across classes: it measures how
spread out each class is around its center.
Their ratio measures separation between classes relative to their internal
variation. The mean class-relative separation ratio increases from
$1.820$ to $2.056$ on ScanNet (19 classes), from $2.854$ to $3.163$
on ScanNet++ (100 classes), and from $2.414$ to $2.738$ on ScanNet++ (20 classes).
The ratio increases in every scene in all three settings.
We conclude that the transformation does not merely spread feature directions more broadly,
it improves their separation relative to the ground-truth semantic classes.

\subsection{Open-Vocabulary Segmentation}\label{sec:eval:ov-seg}
\paragraph{Multi-Class Calibration.}
We present our improvements for multi-class evaluation in the top half of \autoref{tab:calibration-combined}. We report LERF-OVS, 3D-OVS, ScanNet with the 19-class targets and ScanNet++ with the 100-class targets. The smaller ScanNet 10-class and ScanNet++ 20-class vocabularies can be found in \autoref{sec:additional-results}.
The results show that our calibration consistently improves mIoU in all datasets and across all methods, except for VALA and OpenGaussian on LERF-OVS, which are mostly unchanged. We find that our calibration can sometimes shift the precision/recall trade-off compared to the uncalibrated baseline (see \autoref{sec:additional-results}), but in most cases we improve all metrics.

\paragraph{Single-Query Calibration.}
In the bottom half of \autoref{tab:calibration-combined} we present results for our calibration when using only a single query at a time. We query all labels present in a scene individually and average over present classes, then over scenes. 
Unlike the multi-class ablation, these $N=1$ results threshold one query's
target-minus-bank margin rather than taking an argmax over the real class vocabulary.
The raw similarity and calibrated relevance-margin thresholds $\tau$ were determined on held-out scenes and then fixed for all evaluation scenes and queries. The dataset-specific values and LangSplat exceptions are stated in Appendix~\ref{sec:threshold-details}.

The resulting segmentation results are generally below their multi-class counterparts. This pattern is consistent with the absence of mutual exclusion between independently thresholded query masks. As a result different queries could re-use the same primitives that are otherwise competed for in the multi-class case. Nevertheless, our calibration improves IoU in all cases, especially for VALA on LERF-OVS. To investigate why raw performs so poorly in that case, we select an oracle raw-score threshold on each evaluation scene, yielding only 23.71\% mIoU, barely improving over the held-out threshold. It seems that sparse semantic support attenuates rendered scores differently across pixels, affecting absolute thresholding but not class-relative argmax decisions.
Qualitative results are available in \autoref{sec:additional-results}.

\begin{table}[t]
\centering
\captionsetup{font=small}
\caption{\textbf{Segmentation Results Raw vs. Calibrated}. Top: multi-class ($N>1$). Bottom: single-query ($N=1$). The $N=1$ results use the fixed held-out thresholds stated in \autoref{sec:threshold-details}. Smaller ScanNet 10-class and ScanNet++ 20-class results, as well as precision and recall are reported in \autoref{sec:additional-results}.}
\label{tab:calibration-combined}
\vspace{-2mm}
\fontsize{6.5}{8}\selectfont
\setlength{\tabcolsep}{0.6pt}
\renewcommand{\arraystretch}{1.08}
\begin{tabular*}{\textwidth}{@{\extracolsep{\fill}}lcccccccccccc@{}}
\toprule
 & \multicolumn{3}{c}{LERF-OVS} & \multicolumn{3}{c}{3D-OVS} & \multicolumn{3}{c}{ScanNet (19-class)} & \multicolumn{3}{c}{ScanNet++ (100-class)} \\
\cmidrule(lr){2-4}\cmidrule(lr){5-7}\cmidrule(lr){8-10}\cmidrule(lr){11-13}
Method & \multicolumn{1}{c}{Raw} & \multicolumn{1}{c}{Ours} & \multicolumn{1}{c}{$\Delta$} & \multicolumn{1}{c}{Raw} & \multicolumn{1}{c}{Ours} & \multicolumn{1}{c}{$\Delta$} & \multicolumn{1}{c}{Raw} & \multicolumn{1}{c}{Ours} & \multicolumn{1}{c}{$\Delta$} & \multicolumn{1}{c}{Raw} & \multicolumn{1}{c}{Ours} & \multicolumn{1}{c}{$\Delta$} \\
\midrule
\multicolumn{1}{l}{\textbf{Multi-Query} $N>1$} \\[0.35em]
SFS~\cite{xiong2025splatfeaturesolver} & 49.09 & \textbf{65.67} & \textcolor{green!45!black}{$+16.57$} & 70.79 & \textbf{89.80} & \textcolor{green!45!black}{$+19.01$} & 26.87 & \textbf{39.37} & \textcolor{green!45!black}{$+12.50$} & 19.92 & \textbf{23.74} & \textcolor{green!45!black}{$+3.81$} \\
NormLift~\cite{zang2026normlift} & 58.55 & \textbf{60.25} & \textcolor{green!45!black}{$+1.70$} & 70.90 & \textbf{86.11} & \textcolor{green!45!black}{$+15.20$} & 31.40 & \textbf{39.55} & \textcolor{green!45!black}{$+8.14$} & 20.60 & \textbf{24.76} & \textcolor{green!45!black}{$+4.16$} \\
LUDVIG$^\ast$~\cite{marrie2026ludvig} & 38.05 & \textbf{45.44} & \textcolor{green!45!black}{$+7.39$} & 4.15 & \textbf{5.85} & \textcolor{green!45!black}{$+1.70$} & 29.32 & \textbf{39.83} & \textcolor{green!45!black}{$+10.51$} & 20.31 & \textbf{23.96} & \textcolor{green!45!black}{$+3.66$} \\
Occam~\cite{cheng2025occam} & 57.13 & \textbf{61.05} & \textcolor{green!45!black}{$+3.92$} & 66.19 & \textbf{84.21} & \textcolor{green!45!black}{$+18.01$} & 23.87 & \textbf{32.11} & \textcolor{green!45!black}{$+8.24$} & 20.18 & \textbf{23.16} & \textcolor{green!45!black}{$+2.98$} \\
VALA~\cite{wang2026vala} & \textbf{59.83} & 59.68 & \textcolor{red!75!black}{$-0.15$} & 64.16 & \textbf{79.02} & \textcolor{green!45!black}{$+14.86$} & 27.81 & \textbf{36.16} & \textcolor{green!45!black}{$+8.35$} & 19.28 & \textbf{22.90} & \textcolor{green!45!black}{$+3.61$} \\
OpenGaussian~\cite{wu2024opengaussian} & 33.62 & \textbf{46.18} & \textcolor{green!45!black}{$+12.56$} & 23.73 & \textbf{28.81} & \textcolor{green!45!black}{$+5.08$} & 24.34 & \textbf{32.09} & \textcolor{green!45!black}{$+7.76$} & 15.25 & \textbf{17.72} & \textcolor{green!45!black}{$+2.47$} \\
LangSplat~\cite{qin2024langsplat} & 1.49 & \textbf{5.15} & \textcolor{green!45!black}{$+3.66$} & 7.43 & \textbf{15.19} & \textcolor{green!45!black}{$+7.76$} & 3.04 & \textbf{4.00} & \textcolor{green!45!black}{$+0.96$} & 4.50 & \textbf{6.08} & \textcolor{green!45!black}{$+1.58$} \\
\midrule
\multicolumn{13}{l}{\textbf{Single-Query} $N=1$} \\[0.35em]
SFS~\cite{xiong2025splatfeaturesolver} & 45.71 & \textbf{53.86} & \textcolor{green!45!black}{$+8.15$} & 69.21 & \textbf{83.59} & \textcolor{green!45!black}{$+14.39$} & 26.66 & \textbf{34.52} & \textcolor{green!45!black}{$+7.85$} & 22.55 & \textbf{24.39} & \textcolor{green!45!black}{$+1.84$} \\
NormLift~\cite{zang2026normlift} & 48.82 & \textbf{53.03} & \textcolor{green!45!black}{$+4.21$} & 68.62 & \textbf{82.47} & \textcolor{green!45!black}{$+13.86$} & 30.56 & \textbf{36.64} & \textcolor{green!45!black}{$+6.08$} & 23.89 & \textbf{25.54} & \textcolor{green!45!black}{$+1.65$} \\
LUDVIG$^\ast$~\cite{marrie2026ludvig} & 31.98 & \textbf{35.15} & \textcolor{green!45!black}{$+3.16$} & 13.94 & \textbf{13.96} & \textcolor{green!45!black}{$+0.03$} & 28.46 & \textbf{35.76} & \textcolor{green!45!black}{$+7.30$} & 23.11 & \textbf{25.27} & \textcolor{green!45!black}{$+2.16$} \\
Occam~\cite{cheng2025occam} & 46.09 & \textbf{53.67} & \textcolor{green!45!black}{$+7.58$} & 61.10 & \textbf{75.28} & \textcolor{green!45!black}{$+14.17$} & 22.96 & \textbf{31.01} & \textcolor{green!45!black}{$+8.05$} & 22.97 & \textbf{24.24} & \textcolor{green!45!black}{$+1.27$} \\
VALA~\cite{wang2026vala} & 21.30 & \textbf{54.15} & \textcolor{green!45!black}{$+32.85$} & 31.19 & \textbf{75.01} & \textcolor{green!45!black}{$+43.82$} & 26.60 & \textbf{32.85} & \textcolor{green!45!black}{$+6.26$} & 21.99 & \textbf{22.67} & \textcolor{green!45!black}{$+0.68$} \\
OpenGaussian~\cite{wu2024opengaussian} & 30.97 & \textbf{31.06} & \textcolor{green!45!black}{$+0.09$} & 30.72 & \textbf{34.24} & \textcolor{green!45!black}{$+3.52$} & 22.95 & \textbf{30.81} & \textcolor{green!45!black}{$+7.86$} & 18.91 & \textbf{22.08} & \textcolor{green!45!black}{$+3.17$} \\
LangSplat~\cite{qin2024langsplat} & 4.76 & \textbf{5.19} & \textcolor{green!45!black}{$+0.43$} & 14.96 & \textbf{19.57} & \textcolor{green!45!black}{$+4.61$} & 11.37 & \textbf{11.62} & \textcolor{green!45!black}{$+0.25$} & 5.92 & \textbf{8.05} & \textcolor{green!45!black}{$+2.13$} \\
\bottomrule
\multicolumn{13}{l}{\scriptsize $^\ast$ We could not reproduce reasonable 3D-OVS results for LUDVIG.}
\end{tabular*}
\vspace{-3mm}
\end{table}

\subsection{Comparison to other Query-Time Adaptations}
\label{sec:query-comparison}
\begin{wraptable}{r}{0.46\textwidth}
\vspace{-4.6mm}
\centering
\captionof{table}{\textbf{Comparison to other adaptations.} Values are mIoU ($\Delta$ over raw) for $N>1$ on SFS. Other methods are unsuitable for $N=1$.}
\label{tab:external-comparison}
\setlength{\tabcolsep}{1.5pt}
\renewcommand{\arraystretch}{1.08}
\resizebox{\linewidth}{!}{
\begin{tabular}{@{}lcc@{}}
\toprule
Method & \multicolumn{1}{c}{\makebox[6.8em][c]{ScanNet}} & \multicolumn{1}{c}{\makebox[6.8em][c]{ScanNet++}} \\
\midrule
Ours & \makebox[2.7em][r]{$\mathbf{39.37}$}\makebox[4.1em][r]{\textcolor{green!45!black}{$(+12.50)$}} & \makebox[2.7em][r]{$\mathbf{23.74}$}\makebox[4.1em][r]{\textcolor{green!45!black}{$(+3.81)$}} \\
TransCLIP & \makebox[2.7em][r]{$35.54$}\makebox[4.1em][r]{\textcolor{green!45!black}{$(+8.68)$}} & \makebox[2.7em][r]{$22.81$}\makebox[4.1em][r]{\textcolor{green!45!black}{$(+2.89)$}} \\
PTC & \makebox[2.7em][r]{$27.92$}\makebox[4.1em][r]{\textcolor{green!45!black}{$(+1.05)$}} & \makebox[2.7em][r]{$20.81$}\makebox[4.1em][r]{\textcolor{green!45!black}{$(+0.89)$}} \\
Frolic-inspired & \makebox[2.7em][r]{$28.37$}\makebox[4.1em][r]{\textcolor{green!45!black}{$(+1.51)$}} & \makebox[2.7em][r]{$16.90$}\makebox[4.1em][r]{\textcolor{red!75!black}{$(-3.02)$}} \\
UMFC & \makebox[2.7em][r]{$11.31$}\makebox[4.1em][r]{\textcolor{red!75!black}{$(-15.55)$}} & \makebox[2.7em][r]{$9.23$}\makebox[4.1em][r]{\textcolor{red!75!black}{$(-10.69)$}} \\
\bottomrule
\end{tabular}}
\vspace{-4mm}
\end{wraptable}
In \autoref{tab:external-comparison} we compare our method to four other inference-only adaptations using the fixed SFS feature fields. Specifically, we adapt PTC~\citep{ma2026ptc} by replacing
image tokens with valid scene primitives. High-margin initial predictions form
class-specific visual prototypes that interpolate the text embeddings. We apply
this once per scene, without crop-specific processing, and retain the released
defaults ($\rho=0.1$, $K_{\min}=40$, $\mu=0.1$).
Our Frolic-inspired baseline~\citep{zhu2024frolic} estimates covariance from scene
features and text prototypes, using shrinkage of $0.9$. 
It removes the Gaussian intercept and normalizes the
transformed classifier, rather than reproducing full Frolic's fusion and iterative
prior correction. We also adapt UMFC~\citep{liang2024umfc} by clustering valid scene features into six groups and applying the resulting cluster-based corrections to visual features and text prototypes. We retain UMFC's released correction strengths of $0.9$ and $0.3$. Lastly, we adapt TransCLIP~\citep{zanella2024transclip} to each scene
by treating its lifted Gaussian features as unlabeled samples and performing iterative refinement similar to \autoref{sec:method:multiclass}, but instead of the restricted adaptation we use, this baseline employs full TransCLIP with a feature-similarity neighbor graph.

We evaluate these adaptations only for multiclass segmentation, as they require multi-class queries and are inherently unsuited for single-query calibration. The results show that PTC and TransCLIP variants can also improve segmentation performance, while the Frolic variant only improves on ScanNet. Our full calibration outperforms these adaptations, showing that additional feature calibration and reference vocabulary are complementary to the soft clustering idea of TransCLIP.

\subsection{Contribution ablations}
\begin{wraptable}{r}{0.38\textwidth}
\vspace{-4.7mm}
\centering
\captionof{table}{\textbf{Component ablation.} Values are mIoU ($\Delta$ over raw) using SFS.}
\label{tab:component-ablation}
\setlength{\tabcolsep}{1.5pt}
\renewcommand{\arraystretch}{1.08}
\resizebox{\linewidth}{!}{
\begin{tabular}{*{3}{>{\centering\arraybackslash}m{1.8em}}cc}
\toprule
C1 & C2 & C3 & \multicolumn{1}{c}{\makebox[6.8em][c]{ScanNet}} & \multicolumn{1}{c}{\makebox[6.8em][c]{ScanNet++}} \\
\midrule
 &  &  & \makebox[2.7em][r]{$26.87$}\makebox[4.1em][r]{} & \makebox[2.7em][r]{$19.92$}\makebox[4.1em][r]{} \\
$\checkmark$ &  &  & \makebox[2.7em][r]{$31.83$}\makebox[4.1em][r]{\textcolor{green!45!black}{$(+4.97)$}} & \makebox[2.7em][r]{$18.77$}\makebox[4.1em][r]{\textcolor{red!75!black}{$(-1.15)$}} \\
 & $\checkmark$ &  & \makebox[2.7em][r]{$35.09$}\makebox[4.1em][r]{\textcolor{green!45!black}{$(+8.22)$}} & \makebox[2.7em][r]{$23.66$}\makebox[4.1em][r]{\textcolor{green!45!black}{$(+3.73)$}} \\
$\checkmark$ & $\checkmark$ &  & \makebox[2.7em][r]{$36.37$}\makebox[4.1em][r]{\textcolor{green!45!black}{$(+9.50)$}} & \makebox[2.7em][r]{$21.80$}\makebox[4.1em][r]{\textcolor{green!45!black}{$(+1.88)$}} \\
$\checkmark$ &  & $\checkmark$ & \makebox[2.7em][r]{$36.49$}\makebox[4.1em][r]{\textcolor{green!45!black}{$(+9.63)$}} & \makebox[2.7em][r]{$21.15$}\makebox[4.1em][r]{\textcolor{green!45!black}{$(+1.23)$}} \\
$\checkmark$ & $\checkmark$ & $\checkmark$ & \makebox[2.7em][r]{$\mathbf{39.37}$}\makebox[4.1em][r]{\textcolor{green!45!black}{$(+12.50)$}} & \makebox[2.7em][r]{$\mathbf{23.74}$}\makebox[4.1em][r]{\textcolor{green!45!black}{$(+3.81)$}} \\
\bottomrule
\multicolumn{5}{l}{\footnotesize C1 = Feature Calibration, C2 = Score Calibration,}\\
\multicolumn{5}{l}{\footnotesize C3 = Reference Vocabulary}
\end{tabular}}
\end{wraptable}
Table~\ref{tab:component-ablation} evaluates the three conceptual contributions
used by the full $N>1$ method. Feature Calibration combines symmetric centering
and regularized text whitening, Score Calibration uses soft clustering, and Reference Vocabulary enables our bank of generic words. When Reference Vocabulary is turned off, we use the $N>1$ queries as the whitening basis.
The full method yields the largest mIoU gain on both datasets. On ScanNet++ Feature Calibration alone has a negative impact, while being helpful in every combination with other components. In \autoref{app:lambda-alpha} we show additional experiments on sensitivity of $\lambda$ and $\alpha$ parameters.

\section{Conclusion}\label{sec:conclusion}

We studied scene-conditioned calibration for open-vocabulary 3D segmentation while holding reconstructed geometry and lifted features fixed. Combining scene centering, regularized text-prototype whitening, and scene-relative score adjustment improves mIoU across all evaluated lifting methods on four datasets, in both multiclass and single-query settings. Geometric diagnostics show that centering improves class-relative feature separation, while component ablations demonstrate the importance of combining feature, score calibration and our fixed reference vocabulary to determine text-whitening statistics. Multiclass prediction uses soft clustering, while single-query prediction uses a reduced-CSLS margin against the reference vocabulary. These results establish query-time calibration as a useful complement to feature lifting: segmentation can improve without retraining, re-encoding source images, or re-lifting features. 

\paragraph{Limitations.} %
Our evaluation is limited to indoor scenes, and the shared centering and whitening
parameters were selected using SFS results on held-out indoor scenes, leaving transfer to other environments and embedding models untested.
Single-query evaluation uses GT-present classes and held-out-selected thresholds, so it does not establish reliable rejection of entirely absent targets or arbitrary natural-language queries.
Predictions depend on the requested vocabulary and reference bank, whose suitability across different scenes and queries is not established.
Because calibration leaves geometry and feature lifting unchanged, it cannot reconstruct missing surfaces or recover semantic information absent from the lifted field.

\paragraph{Open Questions.}
For single-query segmentation, an especially promising question is how to choose negative queries that reflect the available scene context rather than relying only on a fixed generic bank. Previously queried concepts, when appropriate as negative alternatives, or prior knowledge of the scene could inform the reference vocabulary and provide more relevant competitors for the current target. Growing the reference vocabulary over time as a user queries the scene, especially when applied in similar environments, is an interesting future direction.
Further, all tested scenes are generally in an indoor environment and evaluation in outdoor environments or substantially different scenes remains untested.

\subsection*{AI use statement}

In this work, we used generative AI tools for drafting and editing sections of the paper.
We also used AI tools to find related works beyond what our manual research found. We used AI models to review and critique our paper and used its feedback to improve our work. AI coding agents were used for parts of the implementation and debugging. All generated text, code and suggested papers were reviewed thoroughly by the authors. In all cases we used AI tools as \emph{tools} to assist our work, never to replace or automate full aspects of the research process.

\subsection*{Reproducibility statement}

With our paper we release the full source code required to reproduce our results.
We will upload the reconstructed 3DGS scenes and the lifted feature sets used as input to
our calibration. We will supply a simple Python script that reproduces the exact segmentations that lead to the results reported in Table 1.

\bibliography{iclr2027_conference}
\bibliographystyle{iclr2027_conference}

\appendix
\clearpage
\begin{figure}[t]
\centering
\setlength{\tabcolsep}{0.5mm}
\renewcommand{\arraystretch}{0}
\newcommand{\qualitativerow}[2]{%
  \multicolumn{4}{@{}l@{}}{\rule{0pt}{2.5mm}\scriptsize #1}\\[0.2mm]
  \includegraphics[width=0.245\linewidth]{figures/qualitative/#2/rgb.png} &
  \includegraphics[width=0.245\linewidth]{figures/qualitative/#2/raw.png} &
  \includegraphics[width=0.245\linewidth]{figures/qualitative/#2/calib.png} &
  \includegraphics[width=0.245\linewidth]{figures/qualitative/#2/gt.png}\\[0.6mm]
}
\begin{tabular}{@{}cccc@{}}
\small RGB & \small Raw similarity & \small Ours & \small GT\\[0.6mm]
\qualitativerow{\textbf{Bookshelf} (\texttt{09c1414f1b})}{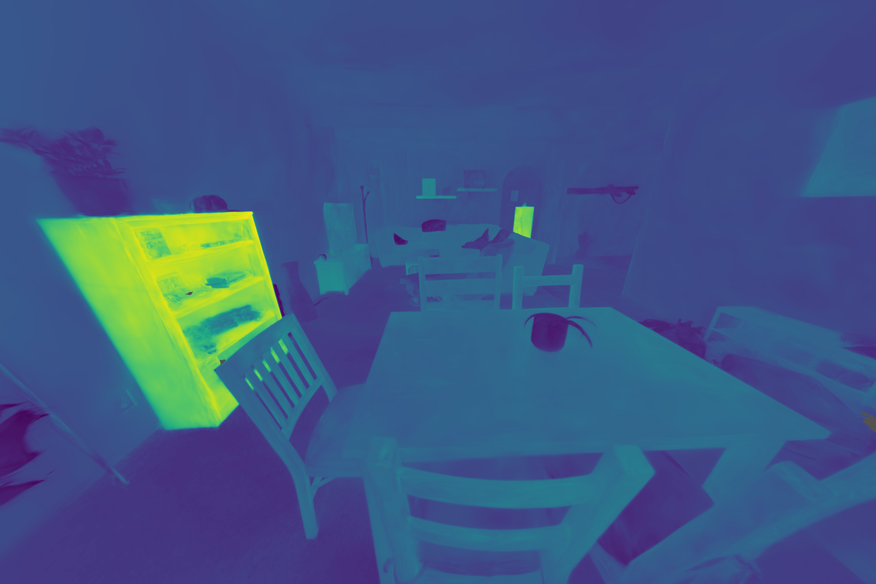}
\qualitativerow{\textbf{Container} (\texttt{3864514494})}{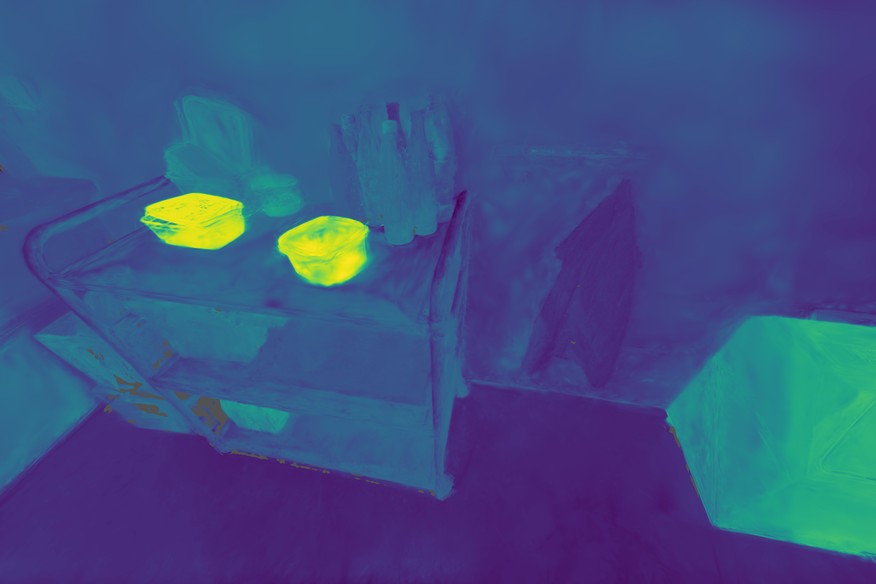}
\qualitativerow{\textbf{Table} (\texttt{3db0a1c8f3})}{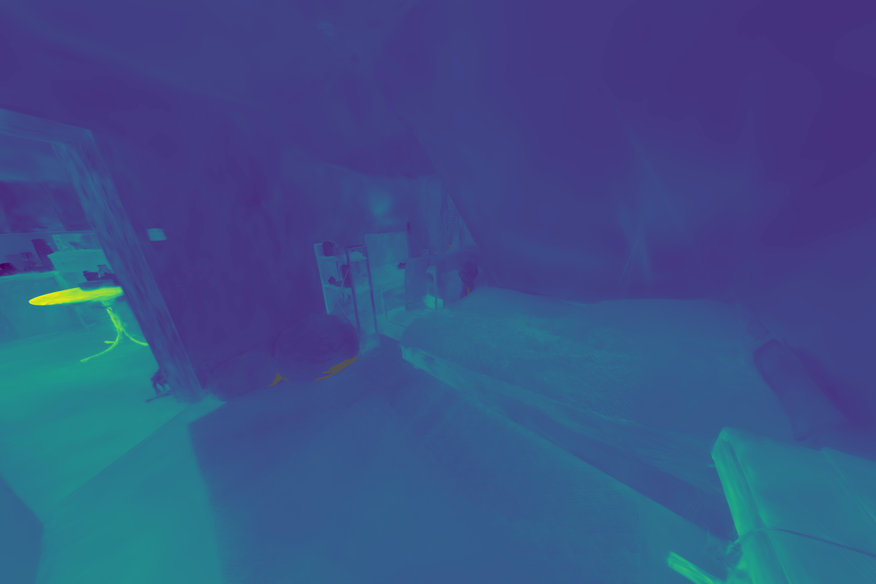}
\qualitativerow{\textbf{Monitor} (\texttt{578511c8a9})}{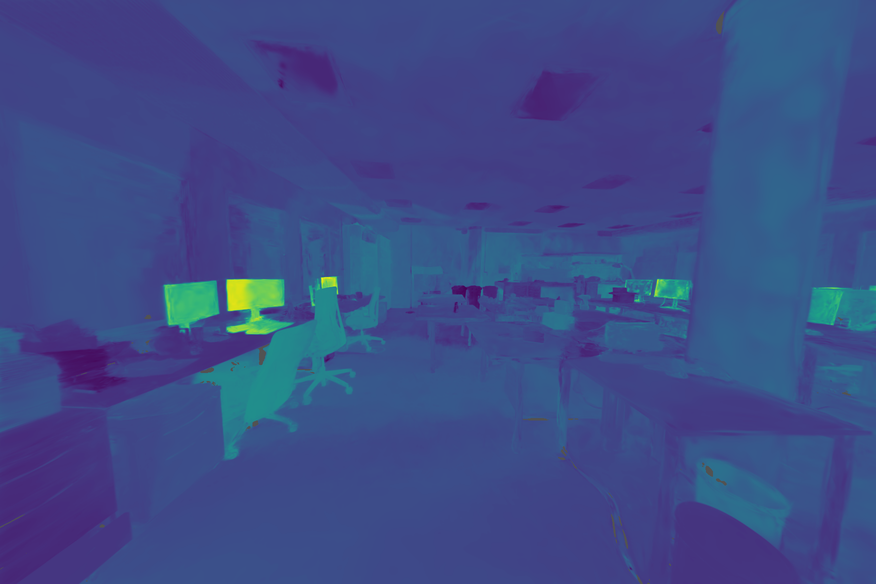}
\qualitativerow{\textbf{Shelf} (\texttt{9071e139d9})}{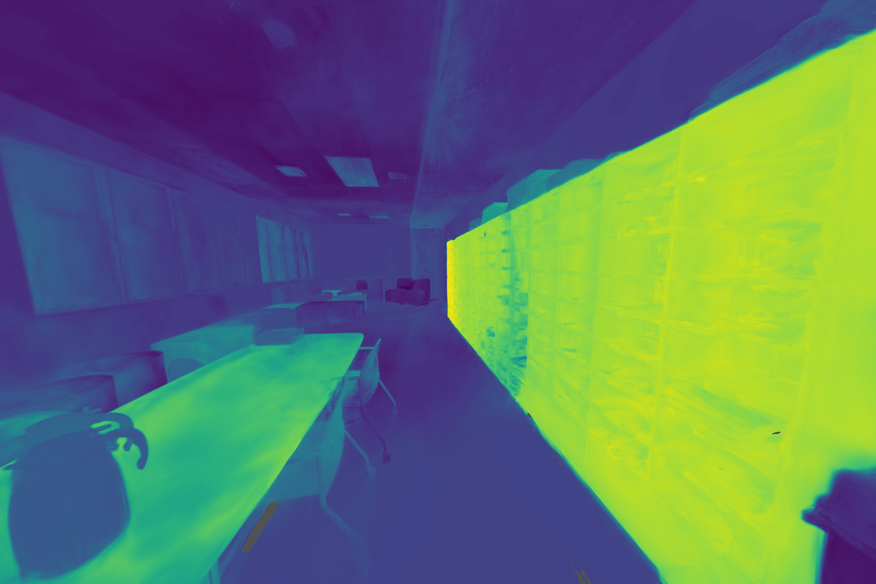}
\qualitativerow{\textbf{Telephone} (\texttt{c50d2d1d42})}{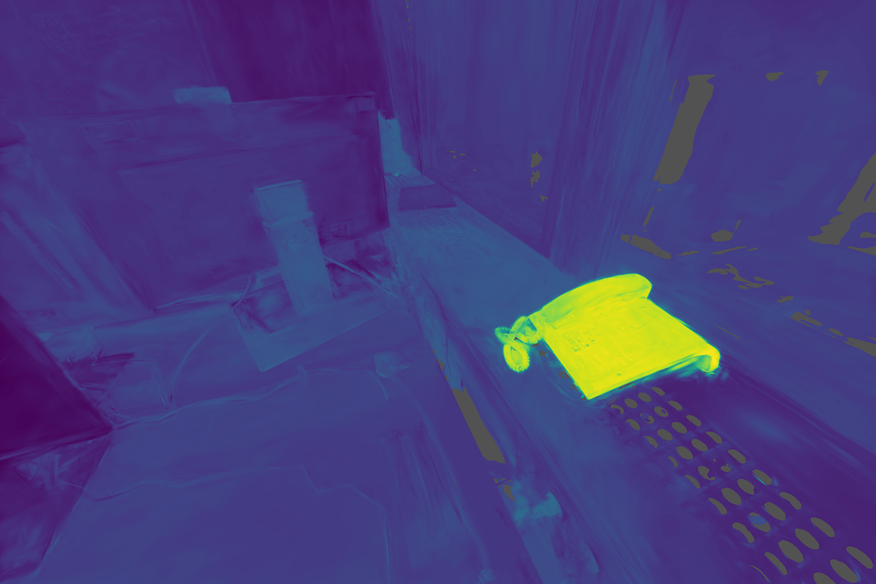}
\qualitativerow{\textbf{Plant} (\texttt{d755b3d9d8})}{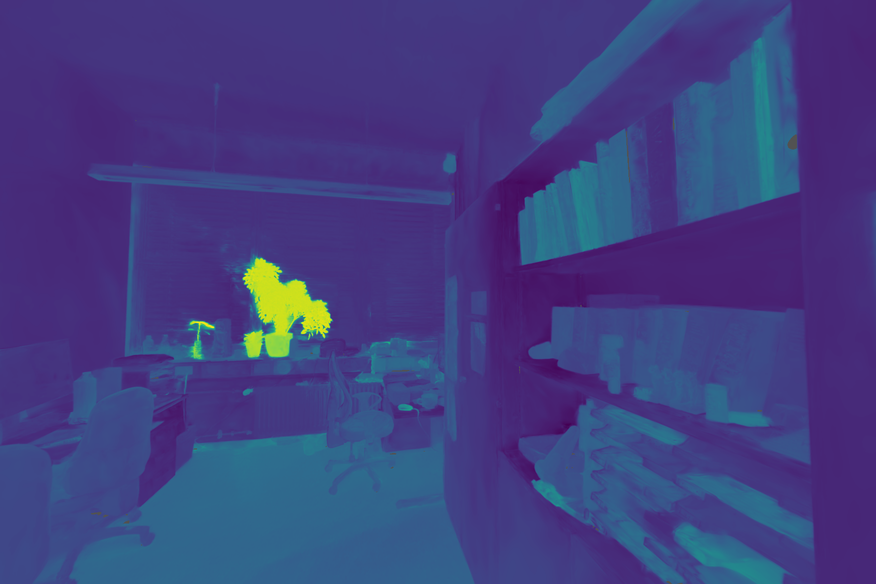}
\qualitativerow{\textbf{Door} (\texttt{e7af285f7d})}{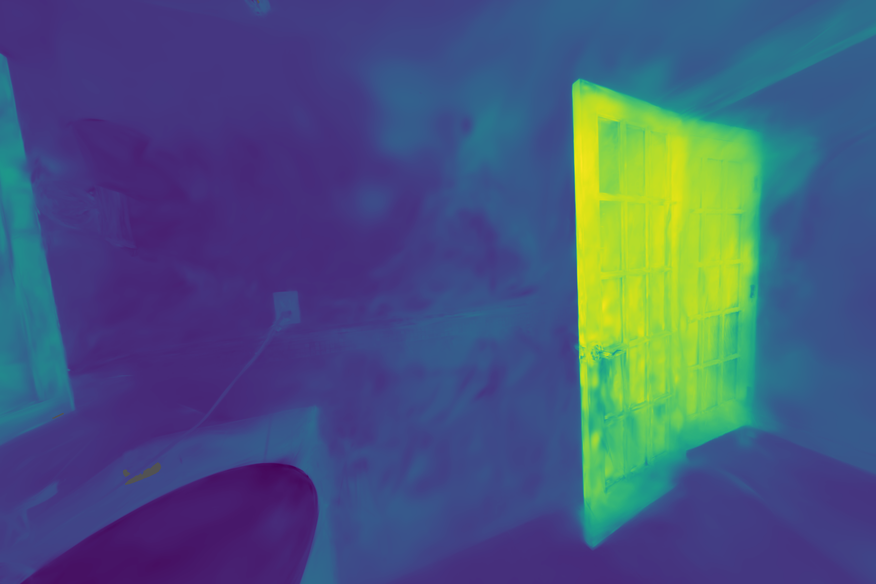}
\end{tabular}
\caption{\textbf{Qualitative query maps on eight ScanNet++ scenes.} Each row shows one query. In mesh-derived binary GT, yellow denotes the query and purple denotes zero. Raw cosine and fully calibrated score maps are independently min--max normalized (purple: low; yellow: high). Note how the calibrated scores increase contrast between foreground and background.}
\label{fig:qualitative-scannetpp}
\end{figure}

\begin{figure}[t]
\centering
\setlength{\tabcolsep}{0.5mm}
\renewcommand{\arraystretch}{0}
\newcommand{\qualitativepanelwidth}{0.205\linewidth}
\newcommand{\quallegenditem}[2]{%
  \mbox{{\color[RGB]{#1}\rule{1.3mm}{1.3mm}}\nobreakspace #2}\hspace{0.6em}%
}
\newcommand{\quallegend}[1]{%
  \par\vspace{0.2mm}
  \parbox{0.96\linewidth}{\scriptsize\centering #1\par}
  \par\vspace{0.7mm}
}
\newcommand{\ovsmulticlassrow}[4]{%
  \multicolumn{4}{@{}l@{}}{\rule{0pt}{2.5mm}\scriptsize #1}\\[0.2mm]
  \includegraphics[width=\qualitativepanelwidth]{figures/#2/rgb.png} &
  \includegraphics[width=\qualitativepanelwidth]{figures/#2/raw_#3.png} &
  \includegraphics[width=\qualitativepanelwidth]{figures/#2/ours_#4.png} &
  \includegraphics[width=\qualitativepanelwidth]{figures/#2/gt.png}\\[0.6mm]
}
\newcommand{\lerfquerycomparison}[6]{%
  \multicolumn{3}{@{}l@{}}{\rule{0pt}{2.5mm}\scriptsize #1}\\[0.2mm]
  \small RGB & \small Raw ($N=1$) & \small Raw ($N>1$)\\[0.6mm]
  \includegraphics[width=\qualitativepanelwidth]{figures/#2/rgb.png} &
  \includegraphics[width=\qualitativepanelwidth]{figures/#2/raw_single_merge_#3.png} &
  \includegraphics[width=\qualitativepanelwidth]{figures/#2/raw_#5.png}\\[0.6mm]
  \small GT & \small Ours ($N=1$) & \small Ours ($N>1$)\\[0.6mm]
  \includegraphics[width=\qualitativepanelwidth]{figures/#2/gt.png} &
  \includegraphics[width=\qualitativepanelwidth]{figures/#2/ours_single_merge_#4.png} &
  \includegraphics[width=\qualitativepanelwidth]{figures/#2/ours_#6.png}\\[0.6mm]
}
\begin{tabular}{@{}cccc@{}}
\small RGB & \small Raw & \small Ours & \small GT\\[0.6mm]
\ovsmulticlassrow{\textbf{3D-OVS: Blue sofa} (\texttt{01\_27})}{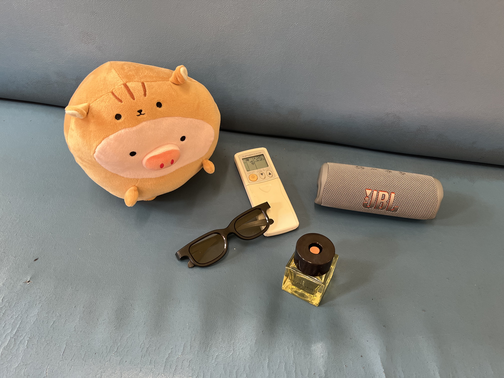}{37}{96}
\end{tabular}
\quallegend{%
\quallegenditem{29,71,113}{a JBL bluetooth speaker}
\quallegenditem{66,168,51}{a bottle of perfume}
\quallegenditem{103,10,244}{a squirrel pig doll}
\quallegenditem{140,107,182}{an aircon controller}
\quallegenditem{177,204,120}{blue-grey sofa}
\quallegenditem{214,46,58}{sunglasses}}
\begin{tabular}{@{}cccc@{}}
\ovsmulticlassrow{\textbf{3D-OVS: Snacks} (\texttt{01\_26})}{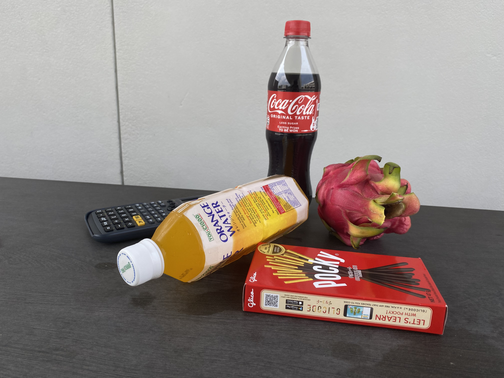}{93}{96}
\end{tabular}
\quallegend{%
\quallegenditem{29,71,113}{Coke Cola}
\quallegenditem{66,168,51}{Glico Pocky chocolate biscuits sticks box}
\quallegenditem{103,10,244}{calculator}
\quallegenditem{140,107,182}{desktop}
\quallegenditem{177,204,120}{orange juice drink}
\quallegenditem{214,46,58}{pitaya}}
\begin{tabular}{@{}cccc@{}}
\ovsmulticlassrow{\textbf{3D-OVS: Table} (\texttt{02\_26})}{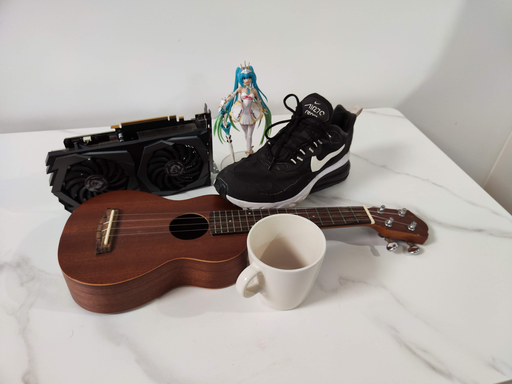}{57}{93}
\end{tabular}
\quallegend{%
\quallegenditem{29,71,113}{a GPU card with fans}
\quallegenditem{66,168,51}{a Hatsune Miku statue}
\quallegenditem{103,10,244}{a beige mug}
\quallegenditem{140,107,182}{a black nike shoe}
\quallegenditem{177,204,120}{a wooden ukulele}
\quallegenditem{214,46,58}{lime wall}}
\begin{tabular}[t]{@{}ccc@{}}
\lerfquerycomparison{\textbf{LERF-OVS: Teatime} (\texttt{02\_frame\_00129})}{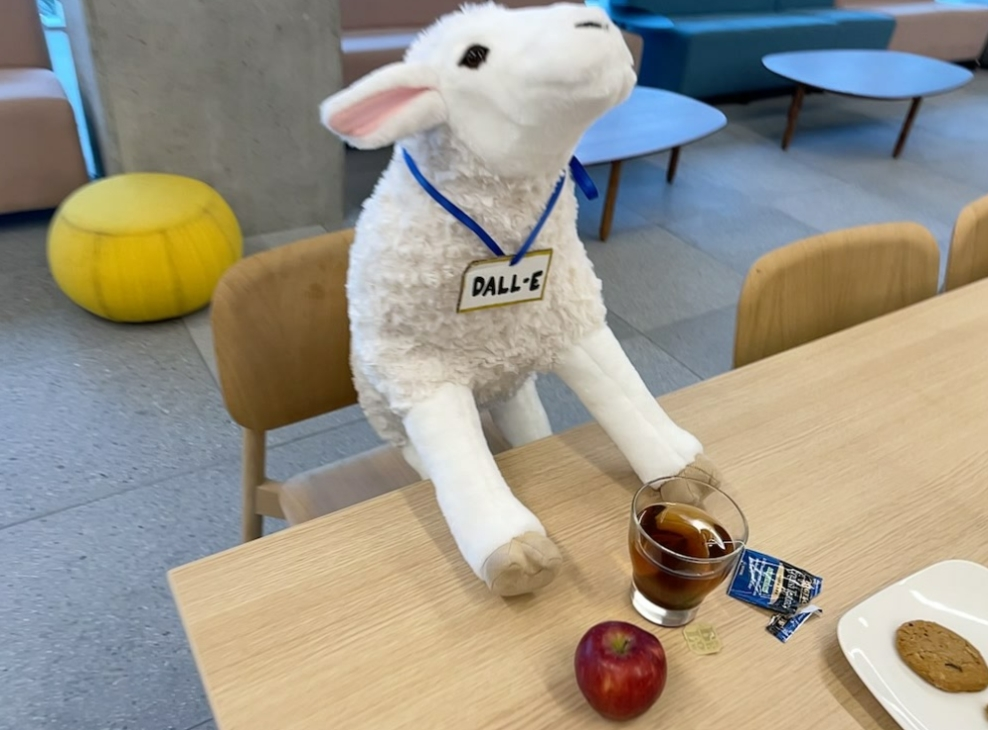}{60}{75}{58}{89}
\end{tabular}\hspace{3mm}%
\parbox[t]{0.28\linewidth}{\scriptsize\raggedright
\vspace{2mm}
\textbf{Classes in GT}\par
\quallegenditem{29,71,113}{apple}\par
\quallegenditem{214,46,58}{dall-e brand}\par
\quallegenditem{70,82,127}{plate}\par
\quallegenditem{107,179,65}{sheep}\par
\quallegenditem{181,118,196}{tea in a glass}\par
\quallegenditem{218,215,134}{three cookies}\par
\quallegenditem{0,57,72}{yellow pouf}\par
\quallegenditem{82,82,82}{unassigned / ignored}\par}
\par\vspace{1mm}
\begin{tabular}[t]{@{}ccc@{}}
\lerfquerycomparison{\textbf{LERF-OVS: Figurines} (\texttt{01\_frame\_00195})}{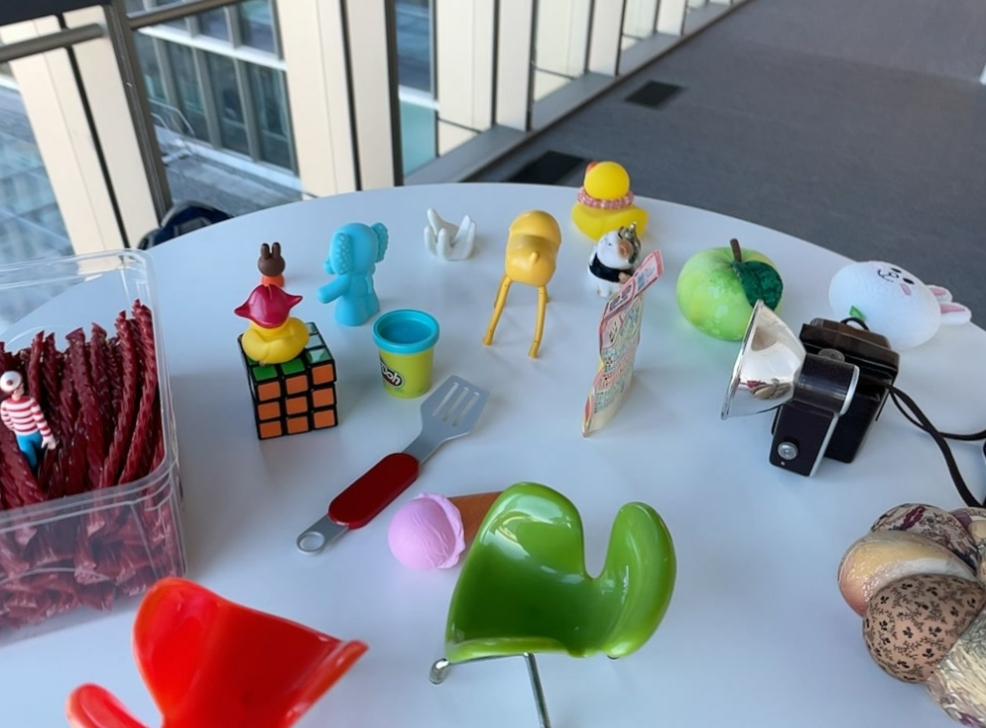}{49}{53}{50}{63}
\end{tabular}\hspace{3mm}%
\parbox[t]{0.28\linewidth}{\scriptsize\raggedright
\vspace{2mm}
\textbf{Classes in GT}\par
\quallegenditem{29,71,113}{bag}\par
\quallegenditem{66,168,51}{green apple}\par
\quallegenditem{103,10,244}{green toy chair}\par
\quallegenditem{177,204,120}{miffy}\par
\quallegenditem{214,46,58}{old camera}\par
\quallegenditem{33,240,189}{pink ice cream}\par
\quallegenditem{70,82,127}{pirate hat}\par
\quallegenditem{107,179,65}{porcelain hand}\par
\quallegenditem{0,57,72}{rubber duck with buoy}\par
\quallegenditem{37,154,10}{rubber duck with hat}\par
\quallegenditem{74,251,203}{rubics cube}\par
\quallegenditem{111,93,141}{spatula}\par
\quallegenditem{185,32,17}{toy cat statue}\par
\quallegenditem{222,129,210}{toy elephant}\par
\quallegenditem{4,226,148}{waldo}\par
\quallegenditem{82,82,82}{unassigned / ignored}\par}
\par\vspace{0.7mm}
\caption{\textbf{Qualitative multiclass segmentation on 3D-OVS and LERF-OVS.} Each 3D-OVS row or LERF-OVS row pair shows one annotated view. For $N>1$, argmax assigns every supported pixel to a queried class, precluding background, merged $N=1$ masks instead retain background where all queries are rejected.}
\label{fig:qualitative-ovs}
\end{figure}

\clearpage
\section{Additional Results}\label{sec:additional-results}

\paragraph{Qualitative results.}
\autoref{fig:qualitative-scannetpp} shows illustrative SFS query maps from eight ScanNet++ scenes, with one query per scene. These examples use the present classes from the 100-class vocabulary and calibration parameters $\lambda=0.45$ and $\alpha=0.6$. The raw and fully calibrated maps are independently min--max normalized for visualization, so they compare spatial response patterns rather than absolute score magnitudes.

\paragraph{More Segmentation results.}
We report detailed results of all our datasets in \autoref{tab:calibration-all-datasets}. The table shows mIoU, precision and recall for all dataset, including the smaller vocabulary versions of ScanNet (10-class) and ScanNet++ (20-class). The LERF-OVS and 3D-OVS numbers use rendered 2D evaluation following their respective protocols.

\paragraph{Single-query component ablations.}
Table~\ref{tab:component-ablation-n1} reports the $N=1$ SFS component ablations with independently optimized thresholds.
\begin{table}[!ht]
\centering
\caption{\textbf{Single-query component ablations.} We report ScanNet and ScanNet++ using SFS in the single-query setting. Each ablations threshold is fit per method and dataset using the held-out calbiration scenes. Both Whiten and CSLS depends on Bank being present, as they are no-ops on a single query. Ablations with Bank use query vs. max-reference margin, otherwise the query score is thresholded directly.}
\label{tab:component-ablation-n1}
\fontsize{7}{9}\selectfont
\setlength{\tabcolsep}{2pt}
\renewcommand{\arraystretch}{1.08}
\begin{tabular*}{\textwidth}{@{\extracolsep{\fill}}ccccccccccc@{}}
\toprule
 & & & & \multicolumn{3}{c}{ScanNet} & \multicolumn{3}{c}{ScanNet++} \\
\cmidrule(lr){5-7}\cmidrule(lr){8-10}
Center & Whiten & CSLS & Bank & mIoU $\uparrow$ & Precision\ $\uparrow$ & Recall\ $\uparrow$ & mIoU $\uparrow$ & Precision\ $\uparrow$ & Recall\ $\uparrow$ \\
\midrule
 & & & & 26.93 & 36.44 & 63.20 & 22.65 & 36.00 & 40.78 \\
$\checkmark$ & & & & 26.40 & 39.05 & 54.18 & 18.60 & 31.22 & 35.32 \\
 & & & $\checkmark$ & 27.47 & 40.41 & 58.59 & 21.71 & 35.96 & 37.87 \\
 & & $\checkmark$ & $\checkmark$ & 28.60 & 40.94 & 60.00 & 24.01 & 36.43 & 41.29 \\
$\checkmark$ & $\checkmark$ & & $\checkmark$ & 30.71 & 47.82 & 57.31 & 20.56 & 34.53 & 38.75 \\
$\checkmark$ & $\checkmark$ & $\checkmark$ & $\checkmark$ & \textbf{34.52} & \textbf{47.87} & \textbf{63.77} & \textbf{24.39} & \textbf{36.14} & \textbf{43.62} \\
\bottomrule
\end{tabular*}
\end{table}

\begin{table}[p]
\centering
\captionsetup{font=small}
\caption{\textbf{Segmentation Results Raw vs. Calibrated across All Datasets.}  Each dataset pair reports multi-query ($N>1$), followed by single-query ($N=1$) results.}
\label{tab:calibration-all-datasets}
\fontsize{6.5}{8}\selectfont
\setlength{\tabcolsep}{0.6pt}
\renewcommand{\arraystretch}{1.08}
\begin{tabular*}{\textwidth}{@{\extracolsep{\fill}}lcccccccccccccccccc@{}}
\bottomrule

\toprule
 & \multicolumn{9}{c}{LERF-OVS} & \multicolumn{9}{c}{3D-OVS} \\
\cmidrule(lr){2-10}\cmidrule(lr){11-19}
& \multicolumn{3}{c}{mIoU $\uparrow$} & \multicolumn{3}{c}{Precision $\uparrow$} & \multicolumn{3}{c}{Recall $\uparrow$} & \multicolumn{3}{c}{mIoU $\uparrow$} & \multicolumn{3}{c}{Precision $\uparrow$} & \multicolumn{3}{c}{Recall $\uparrow$} \\
\cmidrule(lr){2-4}\cmidrule(lr){5-7}\cmidrule(lr){8-10}\cmidrule(lr){11-13}\cmidrule(lr){14-16}\cmidrule(lr){17-19}
Method & \multicolumn{1}{c}{Raw} & \multicolumn{1}{c}{Ours} & \multicolumn{1}{c}{$\Delta$} & \multicolumn{1}{c}{Raw} & \multicolumn{1}{c}{Ours} & \multicolumn{1}{c}{$\Delta$} & \multicolumn{1}{c}{Raw} & \multicolumn{1}{c}{Ours} & \multicolumn{1}{c}{$\Delta$} & \multicolumn{1}{c}{Raw} & \multicolumn{1}{c}{Ours} & \multicolumn{1}{c}{$\Delta$} & \multicolumn{1}{c}{Raw} & \multicolumn{1}{c}{Ours} & \multicolumn{1}{c}{$\Delta$} & \multicolumn{1}{c}{Raw} & \multicolumn{1}{c}{Ours} & \multicolumn{1}{c}{$\Delta$} \\
\midrule
\multicolumn{19}{l}{\textbf{Multi-Query} $N>1$} \\
SFS & 49.09 & \textbf{65.67} & \textcolor{green!45!black}{$+16.57$} & 67.67 & \textbf{77.11} & \textcolor{green!45!black}{$+9.45$} & 61.46 & \textbf{71.23} & \textcolor{green!45!black}{$+9.77$} & 70.79 & \textbf{89.80} & \textcolor{green!45!black}{$+19.01$} & 75.94 & \textbf{94.09} & \textcolor{green!45!black}{$+18.15$} & 88.95 & \textbf{95.58} & \textcolor{green!45!black}{$+6.63$} \\
NormLift & 58.55 & \textbf{60.25} & \textcolor{green!45!black}{$+1.70$} & \textbf{74.58} & 68.42 & \textcolor{red!75!black}{$-6.16$} & \textbf{68.23} & 66.97 & \textcolor{red!75!black}{$-1.26$} & 70.90 & \textbf{86.11} & \textcolor{green!45!black}{$+15.20$} & 74.41 & \textbf{90.49} & \textcolor{green!45!black}{$+16.08$} & 88.00 & \textbf{93.11} & \textcolor{green!45!black}{$+5.10$} \\
LUDVIG & 38.05 & \textbf{45.44} & \textcolor{green!45!black}{$+7.39$} & \textbf{59.77} & 57.75 & \textcolor{red!75!black}{$-2.02$} & 49.76 & \textbf{58.34} & \textcolor{green!45!black}{$+8.57$} & 4.15 & \textbf{5.85} & \textcolor{green!45!black}{$+1.70$} & 10.50 & \textbf{16.80} & \textcolor{green!45!black}{$+6.30$} & 14.07 & \textbf{14.59} & \textcolor{green!45!black}{$+0.52$} \\
Occam & 57.13 & \textbf{61.05} & \textcolor{green!45!black}{$+3.92$} & \textbf{71.66} & 69.71 & \textcolor{red!75!black}{$-1.95$} & 66.00 & \textbf{67.21} & \textcolor{green!45!black}{$+1.21$} & 66.19 & \textbf{84.21} & \textcolor{green!45!black}{$+18.01$} & 73.14 & \textbf{89.04} & \textcolor{green!45!black}{$+15.91$} & 87.94 & \textbf{93.08} & \textcolor{green!45!black}{$+5.14$} \\
VALA & \textbf{59.83} & 59.68 & \textcolor{red!75!black}{$-0.15$} & \textbf{74.89} & 71.72 & \textcolor{red!75!black}{$-3.16$} & \textbf{68.59} & 66.46 & \textcolor{red!75!black}{$-2.12$} & 64.16 & \textbf{79.02} & \textcolor{green!45!black}{$+14.86$} & 71.88 & \textbf{85.49} & \textcolor{green!45!black}{$+13.61$} & 87.23 & \textbf{91.53} & \textcolor{green!45!black}{$+4.30$} \\
OpenGaussian & 33.62 & \textbf{46.18} & \textcolor{green!45!black}{$+12.56$} & 41.57 & \textbf{51.09} & \textcolor{green!45!black}{$+9.51$} & 45.79 & \textbf{55.00} & \textcolor{green!45!black}{$+9.21$} & 23.73 & \textbf{28.81} & \textcolor{green!45!black}{$+5.08$} & 36.45 & \textbf{45.91} & \textcolor{green!45!black}{$+9.46$} & 56.93 & \textbf{59.22} & \textcolor{green!45!black}{$+2.29$} \\
LangSplat & 1.49 & \textbf{5.15} & \textcolor{green!45!black}{$+3.66$} & 15.03 & \textbf{21.44} & \textcolor{green!45!black}{$+6.41$} & 8.88 & \textbf{10.24} & \textcolor{green!45!black}{$+1.37$} & 7.43 & \textbf{15.19} & \textcolor{green!45!black}{$+7.76$} & 26.87 & \textbf{41.79} & \textcolor{green!45!black}{$+14.93$} & 20.38 & \textbf{32.55} & \textcolor{green!45!black}{$+12.17$} \\
\midrule
\multicolumn{19}{l}{\textbf{Single-Query} $N=1$} \\
SFS & 45.71 & \textbf{53.86} & \textcolor{green!45!black}{$+8.15$} & 56.91 & \textbf{67.52} & \textcolor{green!45!black}{$+10.60$} & \textbf{79.10} & 76.00 & \textcolor{red!75!black}{$-3.11$} & 69.21 & \textbf{83.59} & \textcolor{green!45!black}{$+14.39$} & 78.00 & \textbf{90.78} & \textcolor{green!45!black}{$+12.78$} & 82.18 & \textbf{92.01} & \textcolor{green!45!black}{$+9.83$} \\
NormLift & 48.82 & \textbf{53.03} & \textcolor{green!45!black}{$+4.21$} & 63.21 & \textbf{66.89} & \textcolor{green!45!black}{$+3.69$} & \textbf{68.16} & 67.82 & \textcolor{red!75!black}{$-0.34$} & 68.62 & \textbf{82.47} & \textcolor{green!45!black}{$+13.86$} & 77.83 & \textbf{88.63} & \textcolor{green!45!black}{$+10.79$} & 80.08 & \textbf{92.31} & \textcolor{green!45!black}{$+12.23$} \\
LUDVIG & 31.98 & \textbf{35.15} & \textcolor{green!45!black}{$+3.16$} & 42.49 & \textbf{53.14} & \textcolor{green!45!black}{$+10.65$} & \textbf{74.02} & 62.19 & \textcolor{red!75!black}{$-11.83$} & 13.94 & \textbf{13.96} & \textcolor{green!45!black}{$+0.03$} & 16.40 & \textbf{16.42} & \textcolor{green!45!black}{$+0.03$} & 81.98 & \textbf{83.03} & \textcolor{green!45!black}{$+1.05$} \\
Occam & 46.09 & \textbf{53.67} & \textcolor{green!45!black}{$+7.58$} & 64.16 & \textbf{66.38} & \textcolor{green!45!black}{$+2.22$} & 68.07 & \textbf{69.87} & \textcolor{green!45!black}{$+1.80$} & 61.10 & \textbf{75.28} & \textcolor{green!45!black}{$+14.17$} & 77.61 & \textbf{83.39} & \textcolor{green!45!black}{$+5.79$} & 73.99 & \textbf{78.15} & \textcolor{green!45!black}{$+4.16$} \\
VALA & 21.30 & \textbf{54.15} & \textcolor{green!45!black}{$+32.85$} & 42.98 & \textbf{71.03} & \textcolor{green!45!black}{$+28.05$} & 38.65 & \textbf{65.88} & \textcolor{green!45!black}{$+27.22$} & 31.19 & \textbf{75.01} & \textcolor{green!45!black}{$+43.82$} & 54.41 & \textbf{84.59} & \textcolor{green!45!black}{$+30.19$} & 56.48 & \textbf{78.24} & \textcolor{green!45!black}{$+21.76$} \\
OpenGaussian & 30.97 & \textbf{31.06} & \textcolor{green!45!black}{$+0.09$} & 37.35 & \textbf{41.75} & \textcolor{green!45!black}{$+4.40$} & \textbf{56.50} & 41.13 & \textcolor{red!75!black}{$-15.37$} & 30.72 & \textbf{34.24} & \textcolor{green!45!black}{$+3.52$} & 55.69 & \textbf{59.15} & \textcolor{green!45!black}{$+3.46$} & 36.38 & \textbf{36.41} & \textcolor{green!45!black}{$+0.03$} \\
LangSplat & 4.76 & \textbf{5.19} & \textcolor{green!45!black}{$+0.43$} & 8.77 & \textbf{8.89} & \textcolor{green!45!black}{$+0.12$} & \textbf{77.51} & 57.28 & \textcolor{red!75!black}{$-20.23$} & 14.96 & \textbf{19.57} & \textcolor{green!45!black}{$+4.61$} & 17.32 & \textbf{29.49} & \textcolor{green!45!black}{$+12.17$} & \textbf{86.29} & 69.34 & \textcolor{red!75!black}{$-16.95$} \\
\bottomrule

\toprule
 & \multicolumn{9}{c}{ScanNet (10-class)} & \multicolumn{9}{c}{ScanNet (19-class)} \\
\cmidrule(lr){2-10}\cmidrule(lr){11-19}
& \multicolumn{3}{c}{mIoU $\uparrow$} & \multicolumn{3}{c}{Precision $\uparrow$} & \multicolumn{3}{c}{Recall $\uparrow$} & \multicolumn{3}{c}{mIoU $\uparrow$} & \multicolumn{3}{c}{Precision $\uparrow$} & \multicolumn{3}{c}{Recall $\uparrow$} \\
\cmidrule(lr){2-4}\cmidrule(lr){5-7}\cmidrule(lr){8-10}\cmidrule(lr){11-13}\cmidrule(lr){14-16}\cmidrule(lr){17-19}
Method & \multicolumn{1}{c}{Raw} & \multicolumn{1}{c}{Ours} & \multicolumn{1}{c}{$\Delta$} & \multicolumn{1}{c}{Raw} & \multicolumn{1}{c}{Ours} & \multicolumn{1}{c}{$\Delta$} & \multicolumn{1}{c}{Raw} & \multicolumn{1}{c}{Ours} & \multicolumn{1}{c}{$\Delta$} & \multicolumn{1}{c}{Raw} & \multicolumn{1}{c}{Ours} & \multicolumn{1}{c}{$\Delta$} & \multicolumn{1}{c}{Raw} & \multicolumn{1}{c}{Ours} & \multicolumn{1}{c}{$\Delta$} & \multicolumn{1}{c}{Raw} & \multicolumn{1}{c}{Ours} & \multicolumn{1}{c}{$\Delta$} \\
\midrule
\multicolumn{19}{l}{\textbf{Multi-Query} $N>1$} \\
SFS & 38.25 & \textbf{48.91} & \textcolor{green!45!black}{$+10.65$} & 64.23 & \textbf{70.97} & \textcolor{green!45!black}{$+6.74$} & 55.70 & \textbf{65.14} & \textcolor{green!45!black}{$+9.45$} & 26.87 & \textbf{39.37} & \textcolor{green!45!black}{$+12.50$} & 56.38 & \textbf{60.67} & \textcolor{green!45!black}{$+4.30$} & 41.45 & \textbf{51.18} & \textcolor{green!45!black}{$+9.72$} \\
NormLift & 43.89 & \textbf{50.18} & \textcolor{green!45!black}{$+6.29$} & 63.41 & \textbf{71.02} & \textcolor{green!45!black}{$+7.61$} & 63.70 & \textbf{66.14} & \textcolor{green!45!black}{$+2.45$} & 31.40 & \textbf{39.55} & \textcolor{green!45!black}{$+8.14$} & 55.10 & \textbf{60.68} & \textcolor{green!45!black}{$+5.58$} & 48.57 & \textbf{52.64} & \textcolor{green!45!black}{$+4.07$} \\
LUDVIG & 40.70 & \textbf{49.48} & \textcolor{green!45!black}{$+8.78$} & 64.25 & \textbf{71.97} & \textcolor{green!45!black}{$+7.71$} & 59.01 & \textbf{65.10} & \textcolor{green!45!black}{$+6.09$} & 29.32 & \textbf{39.83} & \textcolor{green!45!black}{$+10.51$} & 56.48 & \textbf{60.39} & \textcolor{green!45!black}{$+3.91$} & 44.78 & \textbf{52.81} & \textcolor{green!45!black}{$+8.03$} \\
Occam & 34.74 & \textbf{45.43} & \textcolor{green!45!black}{$+10.69$} & 57.27 & \textbf{67.34} & \textcolor{green!45!black}{$+10.07$} & 53.68 & \textbf{61.50} & \textcolor{green!45!black}{$+7.82$} & 23.87 & \textbf{32.11} & \textcolor{green!45!black}{$+8.24$} & 50.09 & \textbf{55.54} & \textcolor{green!45!black}{$+5.45$} & 39.06 & \textbf{45.25} & \textcolor{green!45!black}{$+6.19$} \\
VALA & 38.94 & \textbf{43.73} & \textcolor{green!45!black}{$+4.79$} & 59.77 & \textbf{66.78} & \textcolor{green!45!black}{$+7.01$} & 58.16 & \textbf{59.42} & \textcolor{green!45!black}{$+1.26$} & 27.81 & \textbf{36.16} & \textcolor{green!45!black}{$+8.35$} & 53.13 & \textbf{57.98} & \textcolor{green!45!black}{$+4.85$} & 43.05 & \textbf{48.98} & \textcolor{green!45!black}{$+5.94$} \\
OpenGaussian & 33.87 & \textbf{41.79} & \textcolor{green!45!black}{$+7.91$} & 54.21 & \textbf{66.08} & \textcolor{green!45!black}{$+11.87$} & 52.33 & \textbf{57.53} & \textcolor{green!45!black}{$+5.19$} & 24.34 & \textbf{32.09} & \textcolor{green!45!black}{$+7.76$} & 46.80 & \textbf{53.16} & \textcolor{green!45!black}{$+6.36$} & 39.68 & \textbf{43.76} & \textcolor{green!45!black}{$+4.08$} \\
LangSplat & 5.97 & \textbf{7.17} & \textcolor{green!45!black}{$+1.20$} & \textbf{30.72} & 29.33 & \textcolor{red!75!black}{$-1.39$} & \textbf{20.82} & 18.44 & \textcolor{red!75!black}{$-2.38$} & 3.04 & \textbf{4.00} & \textcolor{green!45!black}{$+0.96$} & \textbf{19.96} & 18.55 & \textcolor{red!75!black}{$-1.41$} & \textbf{11.34} & 8.06 & \textcolor{red!75!black}{$-3.28$} \\
\midrule
\multicolumn{19}{l}{\textbf{Single-Query} $N=1$} \\
SFS & 30.47 & \textbf{39.91} & \textcolor{green!45!black}{$+9.44$} & 43.91 & \textbf{55.40} & \textcolor{green!45!black}{$+11.50$} & 62.12 & \textbf{65.82} & \textcolor{green!45!black}{$+3.70$} & 26.66 & \textbf{34.52} & \textcolor{green!45!black}{$+7.85$} & 39.77 & \textbf{47.87} & \textcolor{green!45!black}{$+8.10$} & 56.52 & \textbf{63.77} & \textcolor{green!45!black}{$+7.25$} \\
NormLift & 34.56 & \textbf{42.30} & \textcolor{green!45!black}{$+7.74$} & 45.81 & \textbf{57.77} & \textcolor{green!45!black}{$+11.95$} & 66.57 & \textbf{67.20} & \textcolor{green!45!black}{$+0.64$} & 30.56 & \textbf{36.64} & \textcolor{green!45!black}{$+6.08$} & 40.61 & \textbf{49.55} & \textcolor{green!45!black}{$+8.94$} & 64.74 & \textbf{66.94} & \textcolor{green!45!black}{$+2.20$} \\
LUDVIG & 31.70 & \textbf{41.10} & \textcolor{green!45!black}{$+9.39$} & 44.38 & \textbf{56.77} & \textcolor{green!45!black}{$+12.39$} & 62.95 & \textbf{65.95} & \textcolor{green!45!black}{$+3.00$} & 28.46 & \textbf{35.76} & \textcolor{green!45!black}{$+7.30$} & 40.39 & \textbf{49.11} & \textcolor{green!45!black}{$+8.73$} & 59.53 & \textbf{64.45} & \textcolor{green!45!black}{$+4.91$} \\
Occam & 27.66 & \textbf{35.41} & \textcolor{green!45!black}{$+7.75$} & 43.32 & \textbf{52.78} & \textcolor{green!45!black}{$+9.46$} & 50.44 & \textbf{57.58} & \textcolor{green!45!black}{$+7.14$} & 22.96 & \textbf{31.01} & \textcolor{green!45!black}{$+8.05$} & 37.33 & \textbf{46.30} & \textcolor{green!45!black}{$+8.97$} & 44.56 & \textbf{54.45} & \textcolor{green!45!black}{$+9.88$} \\
VALA & 30.26 & \textbf{37.59} & \textcolor{green!45!black}{$+7.33$} & 45.74 & \textbf{55.01} & \textcolor{green!45!black}{$+9.27$} & 53.03 & \textbf{60.58} & \textcolor{green!45!black}{$+7.55$} & 26.60 & \textbf{32.85} & \textcolor{green!45!black}{$+6.26$} & 40.60 & \textbf{47.79} & \textcolor{green!45!black}{$+7.19$} & 48.62 & \textbf{57.65} & \textcolor{green!45!black}{$+9.03$} \\
OpenGaussian & 26.84 & \textbf{35.53} & \textcolor{green!45!black}{$+8.69$} & 40.92 & \textbf{57.90} & \textcolor{green!45!black}{$+16.98$} & 51.75 & \textbf{52.98} & \textcolor{green!45!black}{$+1.23$} & 22.95 & \textbf{30.81} & \textcolor{green!45!black}{$+7.86$} & 36.56 & \textbf{49.63} & \textcolor{green!45!black}{$+13.07$} & 45.94 & \textbf{53.48} & \textcolor{green!45!black}{$+7.54$} \\
LangSplat & 15.02 & \textbf{15.31} & \textcolor{green!45!black}{$+0.29$} & 15.07 & \textbf{16.46} & \textcolor{green!45!black}{$+1.39$} & \textbf{97.86} & 86.32 & \textcolor{red!75!black}{$-11.54$} & 11.37 & \textbf{11.62} & \textcolor{green!45!black}{$+0.25$} & 11.41 & \textbf{12.71} & \textcolor{green!45!black}{$+1.30$} & \textbf{97.91} & 85.59 & \textcolor{red!75!black}{$-12.32$} \\
\bottomrule

\toprule
 & \multicolumn{9}{c}{ScanNet++ (20-class)} & \multicolumn{9}{c}{ScanNet++ (100-class)} \\
\cmidrule(lr){2-10}\cmidrule(lr){11-19}
 & \multicolumn{3}{c}{mIoU $\uparrow$} & \multicolumn{3}{c}{Precision $\uparrow$} & \multicolumn{3}{c}{Recall $\uparrow$} & \multicolumn{3}{c}{mIoU $\uparrow$} & \multicolumn{3}{c}{Precision $\uparrow$} & \multicolumn{3}{c}{Recall $\uparrow$} \\
\cmidrule(lr){2-4}\cmidrule(lr){5-7}\cmidrule(lr){8-10}\cmidrule(lr){11-13}\cmidrule(lr){14-16}\cmidrule(lr){17-19}
Method & \multicolumn{1}{c}{Raw} & \multicolumn{1}{c}{Ours} & \multicolumn{1}{c}{$\Delta$} & \multicolumn{1}{c}{Raw} & \multicolumn{1}{c}{Ours} & \multicolumn{1}{c}{$\Delta$} & \multicolumn{1}{c}{Raw} & \multicolumn{1}{c}{Ours} & \multicolumn{1}{c}{$\Delta$} & \multicolumn{1}{c}{Raw} & \multicolumn{1}{c}{Ours} & \multicolumn{1}{c}{$\Delta$} & \multicolumn{1}{c}{Raw} & \multicolumn{1}{c}{Ours} & \multicolumn{1}{c}{$\Delta$} & \multicolumn{1}{c}{Raw} & \multicolumn{1}{c}{Ours} & \multicolumn{1}{c}{$\Delta$} \\
\midrule
\multicolumn{19}{l}{\textbf{Multi-Query} $N>1$} \\
SFS & 27.70 & \textbf{33.31} & \textcolor{green!45!black}{$+5.61$} & 45.81 & \textbf{52.83} & \textcolor{green!45!black}{$+7.02$} & 43.77 & \textbf{44.79} & \textcolor{green!45!black}{$+1.02$} & 19.92 & \textbf{23.74} & \textcolor{green!45!black}{$+3.81$} & 39.70 & \textbf{41.99} & \textcolor{green!45!black}{$+2.28$} & 28.23 & \textbf{30.83} & \textcolor{green!45!black}{$+2.60$} \\
NormLift & 28.89 & \textbf{34.03} & \textcolor{green!45!black}{$+5.14$} & 46.20 & \textbf{52.76} & \textcolor{green!45!black}{$+6.56$} & 44.76 & \textbf{45.19} & \textcolor{green!45!black}{$+0.43$} & 20.60 & \textbf{24.76} & \textcolor{green!45!black}{$+4.16$} & 39.19 & \textbf{42.05} & \textcolor{green!45!black}{$+2.86$} & 29.70 & \textbf{32.22} & \textcolor{green!45!black}{$+2.52$} \\
LUDVIG & 28.39 & \textbf{34.10} & \textcolor{green!45!black}{$+5.71$} & 47.25 & \textbf{53.87} & \textcolor{green!45!black}{$+6.62$} & 44.42 & \textbf{45.75} & \textcolor{green!45!black}{$+1.33$} & 20.31 & \textbf{23.96} & \textcolor{green!45!black}{$+3.66$} & 40.81 & \textbf{41.24} & \textcolor{green!45!black}{$+0.43$} & 28.54 & \textbf{30.62} & \textcolor{green!45!black}{$+2.07$} \\
Occam & 27.62 & \textbf{33.30} & \textcolor{green!45!black}{$+5.68$} & 44.33 & \textbf{51.57} & \textcolor{green!45!black}{$+7.24$} & 43.85 & \textbf{45.32} & \textcolor{green!45!black}{$+1.46$} & 20.18 & \textbf{23.16} & \textcolor{green!45!black}{$+2.98$} & 39.07 & \textbf{40.35} & \textcolor{green!45!black}{$+1.28$} & 29.34 & \textbf{30.09} & \textcolor{green!45!black}{$+0.75$} \\
VALA & 26.93 & \textbf{31.94} & \textcolor{green!45!black}{$+5.01$} & 44.25 & \textbf{49.80} & \textcolor{green!45!black}{$+5.55$} & 43.01 & \textbf{43.93} & \textcolor{green!45!black}{$+0.92$} & 19.28 & \textbf{22.90} & \textcolor{green!45!black}{$+3.61$} & 38.69 & \textbf{40.75} & \textcolor{green!45!black}{$+2.06$} & 28.27 & \textbf{30.18} & \textcolor{green!45!black}{$+1.92$} \\
OpenGaussian & 24.53 & \textbf{28.36} & \textcolor{green!45!black}{$+3.83$} & 37.53 & \textbf{44.29} & \textcolor{green!45!black}{$+6.76$} & 40.29 & \textbf{40.85} & \textcolor{green!45!black}{$+0.56$} & 15.25 & \textbf{17.72} & \textcolor{green!45!black}{$+2.47$} & 24.18 & \textbf{29.95} & \textcolor{green!45!black}{$+5.76$} & 23.25 & \textbf{24.49} & \textcolor{green!45!black}{$+1.23$} \\
LangSplat & 9.53 & \textbf{11.59} & \textcolor{green!45!black}{$+2.06$} & 20.60 & \textbf{26.80} & \textcolor{green!45!black}{$+6.21$} & \textbf{19.31} & 18.84 & \textcolor{red!75!black}{$-0.47$} & 4.50 & \textbf{6.08} & \textcolor{green!45!black}{$+1.58$} & 11.77 & \textbf{15.80} & \textcolor{green!45!black}{$+4.04$} & 8.15 & \textbf{9.29} & \textcolor{green!45!black}{$+1.13$} \\
\midrule
\multicolumn{19}{l}{\textbf{Single-Query} $N=1$} \\
SFS & 26.64 & \textbf{29.92} & \textcolor{green!45!black}{$+3.28$} & 37.94 & \textbf{46.51} & \textcolor{green!45!black}{$+8.57$} & \textbf{54.36} & 48.55 & \textcolor{red!75!black}{$-5.81$} & 22.55 & \textbf{24.39} & \textcolor{green!45!black}{$+1.84$} & 34.64 & \textbf{36.14} & \textcolor{green!45!black}{$+1.51$} & 42.56 & \textbf{43.62} & \textcolor{green!45!black}{$+1.06$} \\
NormLift & 28.20 & \textbf{31.37} & \textcolor{green!45!black}{$+3.17$} & 39.76 & \textbf{45.65} & \textcolor{green!45!black}{$+5.89$} & \textbf{55.15} & 51.62 & \textcolor{red!75!black}{$-3.53$} & 23.89 & \textbf{25.54} & \textcolor{green!45!black}{$+1.65$} & 35.33 & \textbf{35.39} & \textcolor{green!45!black}{$+0.07$} & 44.01 & \textbf{46.87} & \textcolor{green!45!black}{$+2.87$} \\
LUDVIG & 27.13 & \textbf{30.90} & \textcolor{green!45!black}{$+3.77$} & 37.29 & \textbf{46.84} & \textcolor{green!45!black}{$+9.55$} & \textbf{56.99} & 50.11 & \textcolor{red!75!black}{$-6.89$} & 23.11 & \textbf{25.27} & \textcolor{green!45!black}{$+2.16$} & 34.85 & \textbf{36.83} & \textcolor{green!45!black}{$+1.98$} & 44.67 & \textbf{45.30} & \textcolor{green!45!black}{$+0.63$} \\
Occam & 26.81 & \textbf{30.06} & \textcolor{green!45!black}{$+3.25$} & 38.99 & \textbf{46.34} & \textcolor{green!45!black}{$+7.35$} & \textbf{52.98} & 49.02 & \textcolor{red!75!black}{$-3.96$} & 22.97 & \textbf{24.24} & \textcolor{green!45!black}{$+1.27$} & 35.08 & \textbf{35.21} & \textcolor{green!45!black}{$+0.13$} & 42.66 & \textbf{44.25} & \textcolor{green!45!black}{$+1.59$} \\
VALA & 25.45 & \textbf{27.42} & \textcolor{green!45!black}{$+1.97$} & 39.02 & \textbf{45.85} & \textcolor{green!45!black}{$+6.83$} & \textbf{48.53} & 44.32 & \textcolor{red!75!black}{$-4.22$} & 21.99 & \textbf{22.67} & \textcolor{green!45!black}{$+0.68$} & 35.53 & \textbf{35.61} & \textcolor{green!45!black}{$+0.07$} & 39.08 & \textbf{40.08} & \textcolor{green!45!black}{$+1.00$} \\
OpenGaussian & 24.13 & \textbf{27.99} & \textcolor{green!45!black}{$+3.86$} & 34.16 & \textbf{41.88} & \textcolor{green!45!black}{$+7.72$} & \textbf{50.56} & 50.25 & \textcolor{red!75!black}{$-0.31$} & 18.91 & \textbf{22.08} & \textcolor{green!45!black}{$+3.17$} & 26.33 & \textbf{30.59} & \textcolor{green!45!black}{$+4.26$} & 36.94 & \textbf{43.01} & \textcolor{green!45!black}{$+6.07$} \\
LangSplat & 9.27 & \textbf{11.96} & \textcolor{green!45!black}{$+2.69$} & 16.59 & \textbf{18.79} & \textcolor{green!45!black}{$+2.20$} & 22.81 & \textbf{31.57} & \textcolor{green!45!black}{$+8.76$} & 5.92 & \textbf{8.05} & \textcolor{green!45!black}{$+2.13$} & 11.77 & \textbf{12.83} & \textcolor{green!45!black}{$+1.07$} & 15.65 & \textbf{24.06} & \textcolor{green!45!black}{$+8.41$} \\
\bottomrule

\toprule
\end{tabular*}
\end{table}

\section{Reference Vocabulary Bank}
\label{sec:auxiliary-vocabulary}
\begin{wraptable}{r}{0.45\textwidth}
\vspace{-4mm}
\centering
\caption{\textbf{Reference Vocabulary.}}
\vspace{-3mm}
\resizebox{\linewidth}{!}{
\begin{tabular}{@{}llll@{}}
\toprule
object & stuff & surface & material \\
background & structure & furniture & room \\
clutter & corner & space & interior \\
scene & region & pattern & shape \\
volume & boundary & arrangement & composition \\
portion & section & expanse & form \\
void & span & zone & patch \\
field & bulk & plane & ridge \\
groove & seam & face & side \\
depth & extent & scale & spacing \\
texture detail & contour & profile & margin \\
interval &  &  &  \\
\bottomrule
\end{tabular}}
\vspace{-4mm}
\end{wraptable}
We use bare phrases, without a prompt template. The same bank is used across scenes, datasets, and lifting methods.
Its strings have no exact, case-insensitive match to the evaluated class names, although broad semantic overlap (e.g., ``furniture'') is not excluded.

For whitening, the mean and covariance are fitted once from the 45 words. Requested queries are then transformed by this
fixed reference calibration. Adding, removing, or regrouping queries
does not change the fitted statistics. For $N>1$, bank rows are discarded
before class scoring. For $N=1$, the transformed bank scores remain available
for the target-minus-maximum-bank rejection margin
(Section~\ref{sec:method:single-query}), but the query is not included in the
mean or covariance estimate.

\section{Held-out thresholds}
\label{sec:threshold-details}
For the main $N=1$ evaluations, we use thresholds determined on held-out
scenes and keep them fixed across all evaluation scenes and queries.
On ScanNet, Raw uses $\tau=0.232$ for all methods except LangSplat, which uses $\tau=0.15$.
Ours uses $\tau=-0.040$ for all methods except OpenGaussian, which uses $\tau=-0.024$, and
LangSplat, which uses $\tau=-0.100$.
On ScanNet++, every method uses Raw $\tau=0.245$. Ours uses $\tau=-0.020$
for SFS, LUDVIG, and OpenGaussian, $\tau=-0.022$ for NormLift,
$\tau=-0.017$ for Occam, $\tau=-0.008$ for VALA, and $\tau=-0.040$ for
LangSplat. Raw cosine scores and reference-only
relevance margins have separate thresholds. We apply these fixed values
directly, without refitting thresholds or imposing a recall constraint during
this evaluation.

\section{Sensitivity of $\lambda$ and $\alpha$}
\label{app:lambda-alpha}
\begin{table}[h]
\centering
\caption{\textbf{Parameter sensitivity.} SFS multiclass mIoU (\%) on ScanNet (19 classes) / ScanNet++ (100 classes) is shown on the top / bottom of each cell. The bold cell marks the $(\lambda, \alpha)$ chosen by held-out calibration.}
\label{tab:lambda-alpha-dense}
\setlength{\tabcolsep}{2pt}
\renewcommand{\arraystretch}{1.34}
\begin{tabular}{@{}c*{5}{c}@{}}
\toprule
$\lambda\backslash\alpha$ & 0.30 & 0.35 & 0.40 & 0.45 & 0.50 \\
\midrule
0.20 & \cellcolor[RGB]{100,160,210}\shortstack{40.06\\23.35} & \cellcolor[RGB]{117,170,215}\shortstack{39.95\\23.34} & \cellcolor[RGB]{120,172,216}\shortstack{39.91\\23.36} & \cellcolor[RGB]{111,167,213}\shortstack{39.88\\23.45} & \cellcolor[RGB]{119,172,216}\shortstack{39.86\\23.41} \\
0.25 & \cellcolor[RGB]{108,165,212}\shortstack{39.84\\23.51} & \cellcolor[RGB]{134,181,220}\shortstack{39.73\\23.44} & \cellcolor[RGB]{135,182,220}\shortstack{39.69\\23.47} & \cellcolor[RGB]{120,172,216}\shortstack{39.50\\23.77} & \cellcolor[RGB]{138,183,221}\shortstack{39.41\\23.74} \\
0.30 & \cellcolor[RGB]{157,195,227}\shortstack{39.47\\23.55} & \cellcolor[RGB]{143,187,223}\shortstack{39.40\\23.71} & \cellcolor[RGB]{144,187,223}\shortstack{\textbf{39.37}\\\textbf{23.74}} & \cellcolor[RGB]{152,192,225}\shortstack{39.35\\23.70} & \cellcolor[RGB]{148,190,224}\shortstack{39.38\\23.69} \\
0.35 & \cellcolor[RGB]{212,229,242}\shortstack{39.16\\23.48} & \cellcolor[RGB]{226,237,247}\shortstack{39.14\\23.41} & \cellcolor[RGB]{217,232,244}\shortstack{39.12\\23.49} & \cellcolor[RGB]{250,252,254}\shortstack{38.88\\23.50} & \cellcolor[RGB]{251,252,254}\shortstack{38.87\\23.51} \\
0.40 & \cellcolor[RGB]{133,180,220}\shortstack{39.38\\23.80} & \cellcolor[RGB]{249,251,253}\shortstack{38.74\\23.64} & \cellcolor[RGB]{255,255,255}\shortstack{38.71\\23.64} & \cellcolor[RGB]{249,251,253}\shortstack{38.68\\23.71} & \cellcolor[RGB]{248,251,253}\shortstack{38.69\\23.70} \\
\bottomrule
\end{tabular}
\end{table}

Table~\ref{tab:lambda-alpha-dense} reports SFS multiclass sensitivity of $\lambda$ and $\alpha$ by evaluating neighboring combinations of our selection. 
We selected $(\lambda,\alpha)=(0.30,0.40)$ by averaging multiclass and single-query mIoU using held-out scenes. The table displays the sensitivity of this choice on our test sets with ScanNet on top and ScanNet++ on the bottom. The bold cell marks the selected pair, not the maximum of the displayed multiclass scores.
The selection based on the calibration scenes did not find the highest scoring combination, but generally $\lambda$ and $\alpha$ are robust to slight variations. Our LERF-OVS and OVS-3D evaluations use the same parameters.

\section{Evaluation population and correspondence protocols}\label{app:p3-protocol}
\label{app:point-correspondence-protocol}

Gaussian-based semantic reconstructions differ both in their geometric parameterization and in the rules used to transfer primitive predictions to annotated points. For example, the official OpenGaussian ScanNet configuration~\citep{wu2024opengaussian} fixes the initial point locations and disables densification, permitting evaluation on aligned point rows. Its evaluator also ignores GT labels at rows whose Gaussian opacity is below $0.1$, which can be a substantial amount in some cases. Dr.~Splat~\citep{kim2025drsplat} instead supports unconstrained Gaussian geometry and describes a Gaussian-level evaluation with Mahalanobis-based pseudo-label assignment and significance weighting.

Some prior evaluation protocols discard GT points when their associated predictions are removed by opacity or feature-validity filtering. This makes the evaluated population method-dependent and in many cases drop a substantial amount of points (see \autoref{tab:p1-discarded-points}). Our evaluation instead retains every active GT point and assigns it to an eligible Gaussian using Mahalanobis distance before evaluating the semantic prediction.
Let $\mathcal{E}_m$ contain the exported Gaussians of method $m$ with opacity at least $0.1$ and a valid feature under the method's existing feature-validity rule. For every active GT point $\vx$, we assign
\begin{equation}
 j_m^*(\vx)=\argmin_{j\in\mathcal{E}_m}
 (\vx-\vmu_{mj})^\top\mSigma_{mj}^{-1}(\vx-\vmu_{mj}),
 \qquad \widehat y_m(\vx)=\widehat y_{m,j_m^*(\vx)}.
\end{equation}
In this geometric equation, $\vx,\vmu_{mj}\in\mathbb R^3$ and $\mSigma_{mj}\in\mathbb R^{3\times3}$. These geometric covariances are distinct from the text covariance in Section~\ref{sec:method:whitening}. Here $\mSigma_{mj}$ is constructed from the activated Gaussian scales and rotations. Opacity is used for candidate eligibility, not as an additional factor in this distance. Neither GT class labels nor predicted semantic classes enter the association rule. The same correspondences are used for raw and calibrated predictions. Multiple GT points may select the same Gaussian.

\paragraph{LangSplat field adaptation.}
For LangSplat~\citep{qin2024langsplat}, we use the decoded 512-dimensional features from its level-3 language field as a fixed field for both Raw and Ours. The original rendered-image evaluator forms relevance maps at three levels, applies spatial smoothing, and selects a level for each query and image using the highest peak relevance. We do not reproduce that query-dependent level selection or spatial-smoothing stage as it interferes with out calibrations. The fixed field provides one representation for multiclass refinement and single-query thresholding without introducing an additional level-selection rule. The reported LangSplat results therefore evaluate this fixed-field adaptation rather than the complete original rendered-image pipeline.

\begin{table}[h]
\centering\small
\caption{\textbf{Discarded prediction statistics.} Low opacity, invalid feature, their intersection (Both), and their union (Either) characterize ineligibility as percentages of primitives. GT dropped and GT unsupported report percentages of GT points omitted from evaluation or left without a usable prediction under prior evaluation rules. In contrast, our evaluation retains every active GT point and assigns it to an eligible Gaussian.}
\label{tab:p1-discarded-points}
\resizebox{\textwidth}{!}{%
\begin{tabular}{llrrrrrr}
\toprule
Dataset & Method & Low opacity & Invalid feature & Both & Either & GT dropped & GT unsupported \\
\midrule
ScanNet & SFS & 26.33 & 21.65 & 19.98 & 28.00 & 0.00 & 1.94 \\
ScanNet & NormLift & 26.33 & 21.67 & 19.55 & 28.46 & 0.00 & 2.14 \\
ScanNet & LUDVIG & 26.33 & 19.30 & 17.61 & 28.02 & 0.00 & 1.98 \\
ScanNet & Occam & 28.38 & 0.00 & 0.00 & 28.38 & 0.00 & 0.00 \\
ScanNet & VALA & 0.23 & 0.00 & 0.00 & 0.23 & 0.00 & 0.00 \\
ScanNet & LangSplat & 26.33 & 0.00 & 0.00 & 26.33 & 0.00 & 0.00 \\
\midrule
ScanNet++ & SFS & 30.95 & 10.12 & 10.00 & 31.07 & 28.10 & 28.10 \\
ScanNet++ & NormLift & 30.95 & 6.55 & 6.36 & 31.14 & 28.12 & 28.12 \\
ScanNet++ & LUDVIG & 9.50 & 0.03 & 0.00 & 9.53 & 12.46 & 12.46 \\
ScanNet++ & Occam & 27.76 & 0.00 & 0.00 & 27.76 & 25.73 & 25.73 \\
ScanNet++ & VALA & 0.00 & 0.00 & 0.00 & 0.00 & 0.00 & 0.00 \\
ScanNet++ & OpenGaussian & 42.59 & 22.58 & 12.77 & 52.41 & 34.44 & 34.44 \\
ScanNet++ & LangSplat & 30.95 & 0.00 & 0.00 & 30.95 & 28.00 & 28.00 \\
\bottomrule
\end{tabular}}
\end{table}

\end{document}